\newif\ifENG
\ENGtrue

\newif\ifQED

\newif\ifTWOC

\newif\ifSUB
\SUBtrue

\newif\ifSAM

\newif\ifCMP
\CMPtrue

\ifENG
    \ifTWOC
        \documentclass[12pt,a4paper,twocolumn]{report}
    \else
        \documentclass[12pt,a4paper]{report}
    \fi
\else
    \ifTWOC
        \documentclass[12pt,a4paper,twocolumn]{jreport}
    \else
        \documentclass[12pt,a4paper]{jreport}
    \fi
\fi

\ifCMP
\fi

\usepackage{theorem}
\theorembodyfont{\normalfont}

\makeatletter
\def\QED{{\unskip\nobreak\hfil\penalty50
\hskip1em\hbox{}\nobreak\hfil $\Box$\parfillskip\z@
\finalhyphendemerits\z@\par}}
\makeatother

\ifENG
\def\Definition{Definition}
\else
\def\Definition{定義}
\fi

\ifQED
\newtheorem{defi}{\Definition}
\newenvironment{definition}{\begin{defi}}{\QED\end{defi}}
\else

\fi

\ifENG
\def\Theorem{Theorem}
\else
\def\Theorem{定理}
\fi

\ifQED
\newtheorem{theo}{\Theorem}
\newenvironment{theorem}{\begin{theo}}{\QED\end{theo}}
\else

\fi

\ifENG
\def\Proposition{Proposition}
\else
\def\Proposition{命題}
\fi

\ifQED
\newtheorem{prop}{\Proposition}
\newenvironment{proposition}{\begin{prop}}{\QED\end{prop}}
\else

\fi

\usepackage[utf8]{inputenc}

\usepackage{amsmath,amssymb,graphicx}
\usepackage{algorithmic,amsfonts,cite,graphicx,textcomp,ulem}
\usepackage{xcolor,tabularx}
\usepackage{datetime}

\usepackage{dtk-logos}
\usepackage{listings}
\usepackage{textcomp}
\usepackage{setspace}
\begin{document}

\begin{titlepage}
    \begin{center}
        \Large 2025 Year \, Master Thesis \\

        \vfill
        \ifSUB
            \huge\textbf{Integrated YOLOP Perception and Lyapunov-based Control for Autonomous Mobile Robot Navigation on Track}\\
        \fi

        \vfill
        \Large Graduate School of Science and Technology \\
         \medskip
        \Large Master’s Program in Science and Technology \\
        \medskip
        \Large Green Science and Engineering Division \\
        \vspace{\baselineskip}
        \Large \textbf{B2378474} \\
        \smallskip
        \Large \textbf{MO CHEN} \\
        \vspace{\baselineskip}
        \large Supervisor \quad Wenjing Cao\\

        \vspace{3\baselineskip}
        \newdateformat{myformat}{\THEYEAR-\twodigit{\THEMONTH}-\twodigit{\THEDAY}}
\myformat\today
    \end{center}

    \ifSAM
        \footnotetext[1]{Sample Mode.}
        \footnotetext[2]{Copyright 2022-2023 Leo Liu. All Rights Reserved.}
    \fi
\end{titlepage}

\cleardoublepage
\pagenumbering{gobble}
\tableofcontents
\cleardoublepage
\pagenumbering{arabic}

\chapter{Introduction}

\section{Research background}
In the 1990s, the modern scientific and technological revolution marked by computer technology, microelectronics technology, information technology, network technology, etc., entered a rapid development stage, which became the intrinsic driving force to promote the development of robotics technology, and robotics technology has developed rapidly. Among them, autonomous mobile robots(AMRs) can rely on the sensors they carry to perceive and understand the external environment, make real-time decisions according to the needs of the task, carry out closed-loop control, and operate in an autonomous or semi-autonomous manner. It is a new type of robot with certain self-learning and adaptive ability in known or unknown environment. Navigation is an important problem that needs to be solved for AMRs to realize autonomous control, which refers to the process of mobile robot sensing the environment and its own state through sensors and learning, and realizing the process of pointing to the target autonomous movement in an obstructed environment. Since the first mobile robot, Shakey, was introduced in the 1960s, mobile robot navigation has been receiving a lot of attention due to its comprehensiveness and practicality \cite{verginis2021adaptive}. As one of the core technologies of robots, navigation provides the basis for robots to accomplish various complex tasks, and makes robots widely used in transportation, cleaning, logistics and other scenarios, and active in production and life.

The framework of mobile robot navigation is a hierarchical structure, as demonstrated in Figure 1.1. This hierarchical framework typically encompasses four key modules: perception, decision, path planning, and robot control \cite{teng2023motion}, and contains the main parts such as mapping \cite{taketomi2017visual} and localization \cite{malagon2015mobile}. The perception module serves as the front end of the mobile robot navigation, utilizing sensors to gather data about the surrounding environment. This data is utilized to perform essential tasks such as localization, obstacle detection, path prediction, object detection, and object tracking. This perception module integrates various data sources like high-definition (HD) maps, cameras, radars, lasers, the global positioning system (GPS), and the inertial measurement unit (IMU) to perform tasks like location pinpointing, object detection, and tracking. These tasks enable the system to identify and comprehend diverse elements and obstacles, including roads, lane lines, people, precise robot locations and predicted trajectories. The decision module receives information from the perception module, and then analyzes and reasons based on this information to make appropriate decisions. It considers factors such as rules of various specific scenarios and needs of service recipients and generates a behavioural strategy to respond to the current scenario. The planning module will devise a collision-free trajectory considering the kinematics and dynamics constraints of the robot. It takes the behavioural strategy provided by the decision module and combines it with priori data and real-time perception data to generate a planned path. This path guides the robot’s driving direction, turns, and ensures the robot reaches its destination safely and efficiently. The control module is responsible for precisely and smoothly tracking the planned path through actual robot operations, including acceleration, steering, etc \cite{pendleton2017perception}. 

\begin{figure}[h]
    \begin{center}
        \includegraphics[width=0.8\textwidth]{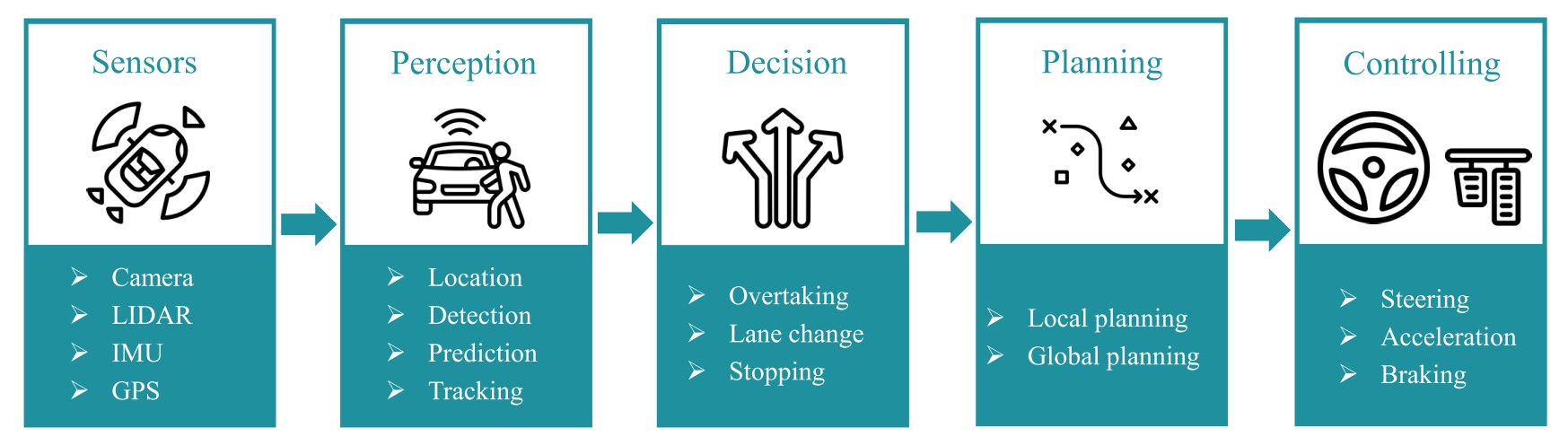}
        \caption{Modules of the autonomous navigation system}
        \label{Figure_1}
    \end{center}
\end{figure}

In order to realize the navigation of AMRs, the perception module recognizes various environmental information through various sensor information: such as road boundaries, terrain features, obstacles, etc. In the navigation of unmanned vehicles, it is also necessary to recognize traffic signs, typical intersections and other information. The robot determines the reachable and unreachable areas in the direction of travel, determines the relative position in the environment, and predicts the movement of dynamic obstacles through environmental perception, thus providing a basis for local path planning. On the simultaneous localization and mapping (SLAM) side, some groups now embed ORB-SLAM3 directly in service-robot pipelines: for example, Xiao et al. enhanced the original front-end with adaptive thresholds and inertial cues to keep hospital logistics robots reliable under extreme illumination swings \cite{xiao2025improving}. Complementing these single-SLAM efforts, the fusion trend remains strong: Yu et al. tightly coupled VINS-Fusion with vehicle-motion constraints to upgrade pose accuracy in autonomous cars by an order of magnitude \cite{yu2022tightly}. Parallel progress in road-structure perception shows a similar pattern.  Kunchala et al. employed LaneNet as the drivable-space extractor in a lightweight navigation module released just last week, demonstrating real-time performance on rural Indian highways \cite{siddaiyan2025enhancing}. Dong et al. blended SCNN layers into a hybrid spatial–temporal encoder-decoder, showing that SCNN-based message passing lifts detection F1 by 4 \% in adverse lighting \cite{dong2023hybrid}. These advancements continue to push the limits of AMR perception, leading to more robust and reliable systems.

The path planning module occupies a central role within the hierarchical framework, acting as a crucial interface between the perception, decision-making, and control modules. Fundamentally, it addresses constrained optimization problems within a complex convex space. In addition, it plays a significant role in multi-agent clustering, obstacle avoidance, and target tracking control, representing a foundational and widely applicable challenge. Therefore, the path planning algorithm constitutes the core component of AMR navigation. Recent progress in local planning for robot navigation has been characterized by algorithm-specific innovations. For conventional planning and control algorithms, Melchiorre et al. introduced a novel APF method enhanced with local attractors to mitigate collision risks and avoid local minima, validated through real-world robotic experiments that demonstrated improved path predictability and obstacle avoidance efficiency \cite{melchiorre2023experiments}. Bui et al. developed an model predictive control (MPC) based local trajectory planner for UAVs, which processes local point cloud data to generate and optimize safe, smooth, and dynamically feasible paths. Simulation results indicated shorter, smoother trajectories and improved energy efficiency \cite{bui2024model}. For reinforcement learning algorithms, to improve learning efficiency, stability, and obstacle avoidance in Gazebo simulations, Chen et al. \cite{chen2024enhanced} proposed an enhanced deep Q-network (DQN) algorithm for indoor mobile robot local path planning, featuring a dueling DQN structure, prioritized experience replay, and maximum entropy integration. Recent advances in robot navigation have yielded increasingly robust and adaptive local planners through algorithm-specific enhancements in control theory and deep reinforcement learning.

\section{Research object}
This study takes an AMR, AI Formula, provided by Honda as the research object. It is a sophisticated system designed with a focus on advanced hardware components to support autonomous driving \cite{published_papers/49092462}. It features a three-wheel configuration, with two differential drive wheels and a passive wheel as illustrated in Figure 1.2. This setup ensures stability, particularly during straight-line driving, by fixing the yaw axis and allowing controlled yaw angles during turns to prevent spinning. The vehicle’s propulsion system includes in-wheel motors with differential drive and a braking system consisting of both disc brakes and regenerative braking. The passive wheel utilizes a spring and damper system for smooth operation, with caster angles and dampers to reduce vibration and enhance handling.

\begin{figure}[h]
    \begin{center}
        \includegraphics[width=0.5\textwidth]{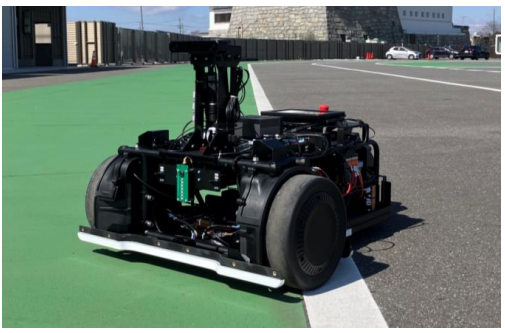}
        \caption{AI Formula robot}
        \label{Figure_2}
    \end{center}
\end{figure}

The robot is equipped with an advanced sensor suite. The stereo camera acquires scene images through synchronized left and right perspectives and uses parallax to calculate depth information, providing the robot with high-precision three-dimensional perception, supporting tasks such as obstacle avoidance, positioning and grasping, and achieving safer, more intelligent environmental interaction and efficient operation capabilities. The inertial measurement unit (IMU) integrates accelerometers, gyroscopes and sometimes magnetometers, and can output 6 DOF motion posture and acceleration information in real time, providing a high-frequency, high-precision data foundation for robot posture solution, stable control and navigation. The wheel encoder detects the wheel rotation angular displacement or pulse number, and outputs the wheel speed and mileage in real time, providing key basic data for the robot chassis closed-loop speed control, odometer positioning and slip detection. It is small in size, fast in response, low in cost, easy to install and maintain, and has strong anti-interference ability. Global Navigation Satellite System (GNSS) is a global autonomous geographic positioning satellite system. It receives multi-constellation satellite signals in real time and outputs centimeter-level positioning, speed, and timing information, providing accurate and reliable basic reference for outdoor robots to build global coordinates, path planning and multi-sensor fusion positioning. Figure 1.3 shows the specific installation locations of these sensors.

\begin{figure}[h]
    \begin{center}
        \includegraphics[width=0.8\textwidth]{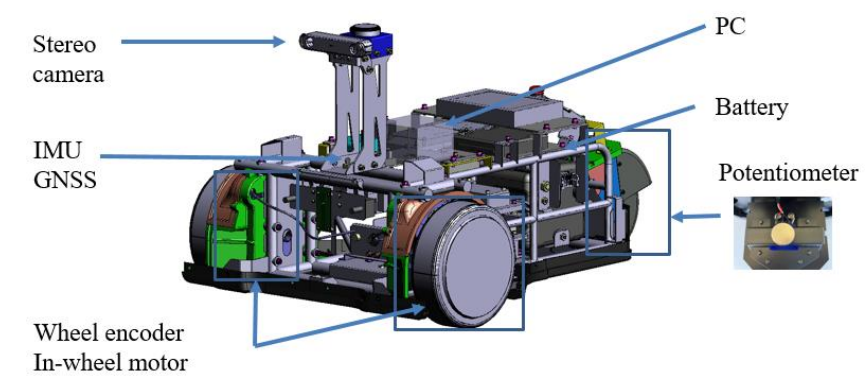}
        \caption{AI Formula robot constituent parts}
        \label{Figure_3}
    \end{center}
\end{figure}

\section{Research purpose}
This study aims to leverage a robot developed by Honda to achieve autonomous navigation on a Formula One race track without relying on HD maps or GNSS data for positioning or path referencing. The top view of the Formula One track is shown in the figure. In this research, a stereo camera is employed to detect and extract lane line features from the environment in real time. These detected lane lines are subsequently used to infer the track’s centerline, which serves as the basis for generating a smooth and continuous navigation path. To ensure that the robot follows this planned path accurately and efficiently, a control algorithm is designed and implemented. The controller enables the robot to perform robust trajectory tracking, allowing it to navigate the track autonomously, smoothly, and at relatively high speeds. This approach emphasizes real-world applicability by minimizing reliance on external infrastructure and prior map data.

\begin{figure}[h]
    \begin{center}
        \includegraphics[width=0.75\textwidth]{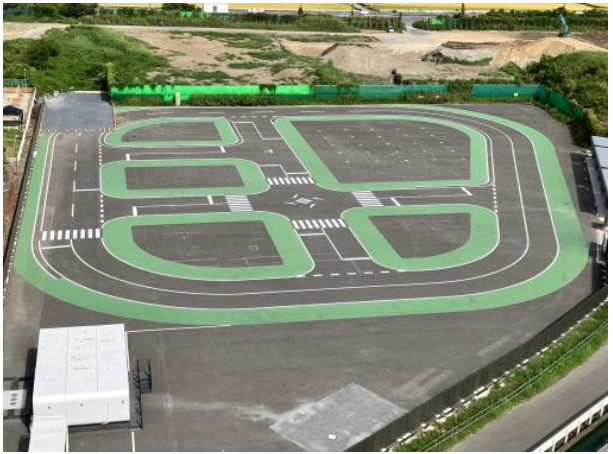}
        \caption{AI Formula race course}
        \label{Figure_4}
    \end{center}
\end{figure}

\section{Thesis Organization}
Chapter 1 introduces the background, object and purposes of the study. Chapter 2, Theoretical Background, serves as a basic introduction and introduces the basic theories and principles related to the study. Chapter 3, Problem Statement, elaborates on the kinematic modeling of the robot and geometric paths. Chapter 4, Control Algorithm discusses the development methods and strategies of the control method in detail. Chapter 5, Results and Discussion, presents the results of the field experiments and provides a comprehensive analysis of the results. Finally, Chapter 6, Conclusion, summarizes this study and proposes directions for future research.
\chapter{Basic Theory and Methods}

\section{YOLOP}
The YOLO (You Only Look Once) series represents a family of real-time object detection algorithms that have significantly advanced the field of computer vision since its inception by Redmon et al. \cite{redmon2016you}. Unlike traditional two-stage detectors such as R-CNN, which separate the processes of region proposal and classification, YOLO treats object detection as a single regression problem, directly predicting bounding boxes and class probabilities from full images in one evaluation. This design dramatically improves inference speed while maintaining competitive accuracy, making YOLO particularly suitable for real-time applications. Over the years, the YOLO architecture has undergone several iterations, each improving upon its predecessor in terms of accuracy, speed, and efficiency. YOLOv3 introduced multi-scale predictions and residual connections, enhancing detection of small objects. YOLOv4 and YOLOv5 further improved performance through better data augmentation, backbone networks, and training strategies. More recent developments, such as YOLOv6 and YOLOv7, have focused on industrial deployment and lightweight design, offering flexible trade-offs between speed and accuracy for various use cases \cite{bochkovskiy2020yolov4}.

YOLOP (You Only Look Once for Panoptic Driving Perception) is a task-specific extension of the YOLO framework designed to address the complex perceptual needs of autonomous driving systems proposed by Wu et al. Unlike conventional approaches that employ separate models for each task, YOLOP achieves end-to-end learning by sharing a common backbone and utilizing task-specific heads, thereby reducing computational cost and improving inference efficiency. As  shown  in Fig 2.1, panoptic  driving  perception  single-shot  network,  YOLOP),  contains  one  shared  encoder  and three  subsequent  decoders  to  solve  specific  tasks. The model is built upon a lightweight variant of YOLOv5, incorporating attention modules and feature pyramid networks to facilitate effective multi-scale feature sharing across tasks. Experimental results on the BDD100K dataset demonstrate that YOLOP achieves real-time performance while maintaining high accuracy across all tasks, highlighting its practicality for real-world deployment in autonomous vehicles \cite{wu2022yolop}. YOLOP represents a significant step toward holistic scene understanding by harmonizing object-level and pixel-level perception within a single framework.

\begin{figure}[h]
    \begin{center}
        \includegraphics[width=0.75\textwidth]{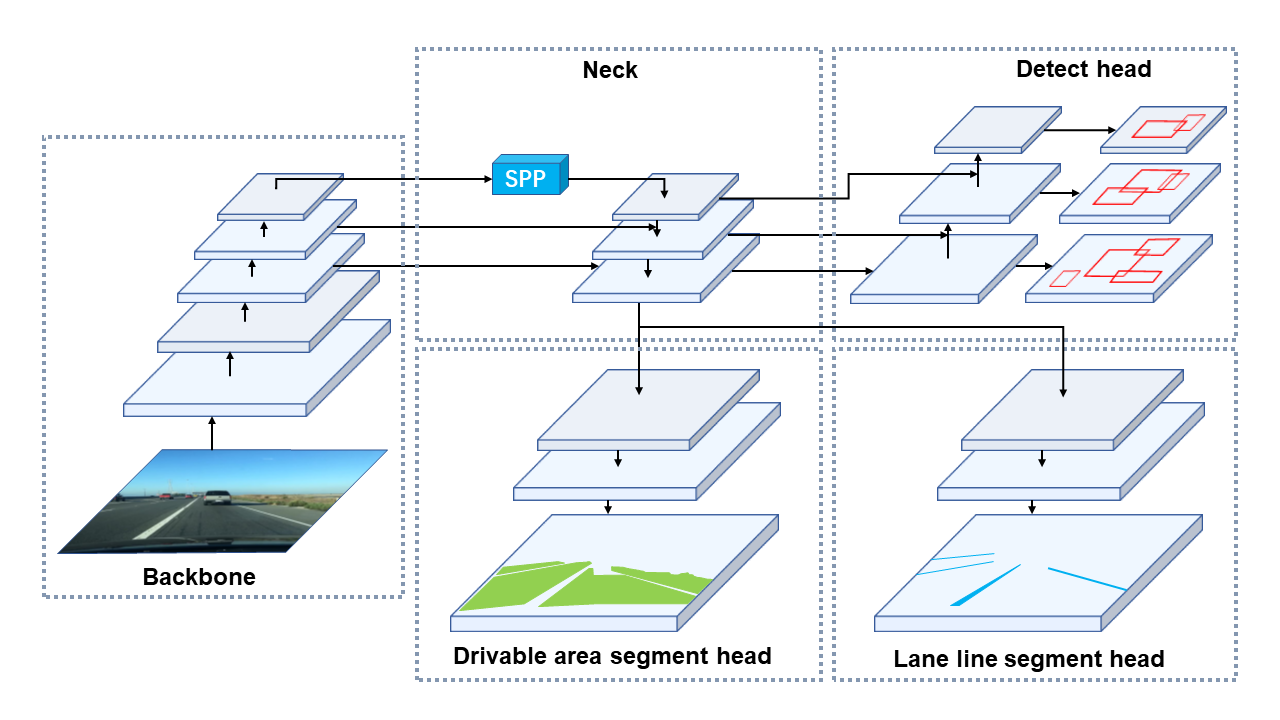}
        \caption{The architecture of YOLOP \cite{wu2022yolop}}
        \label{Figure_4}
    \end{center}
\end{figure}

YOLOP integrates three key perception tasks—object detection, drivable area segmentation, and lane line detection—into a unified multi-task network. Figure 3.1 shows examples of training results for three perception tasks of YOLOP. Among them, lane line detection is formulated as a binary semantic segmentation task, where each pixel in the input image is classified as either belonging to a lane line or the background. This is achieved through a dedicated segmentation head that shares a common encoder with the other tasks. The decoder structure for lane detection involves feeding the low-level feature map from the Feature Pyramid Network (FPN) into a lightweight upsampling path. This path performs three successive upsampling operations to restore the feature map to the original image resolution \( (W, H, 2) \), producing a two-channel output that encodes the per-pixel probability distribution over the lane line and background classes. The loss function for lane detection, denoted as \( L_{\text{ll-seg}} \), combines a standard cross-entropy loss \( L_{\text{ce}} \) with an Intersection-over-Union (IoU) loss \( L_{\text{IoU}} \), as shown in the following equation:
\begin{equation}
L_{\text{ll-seg}} = L_{\text{ce}} + L_{\text{IoU}}.
\end{equation}
The inclusion of IoU loss is particularly important for lane line detection, as lane markings are typically sparse and thin. IoU loss provides stronger supervision for such classes by explicitly penalizing mismatches between the predicted and ground truth segmentation regions. This dual-loss formulation improves the model’s ability to localize narrow structures like lane lines with higher accuracy.

\begin{figure}[h]
    \begin{center}
        \includegraphics[width=0.75\textwidth]{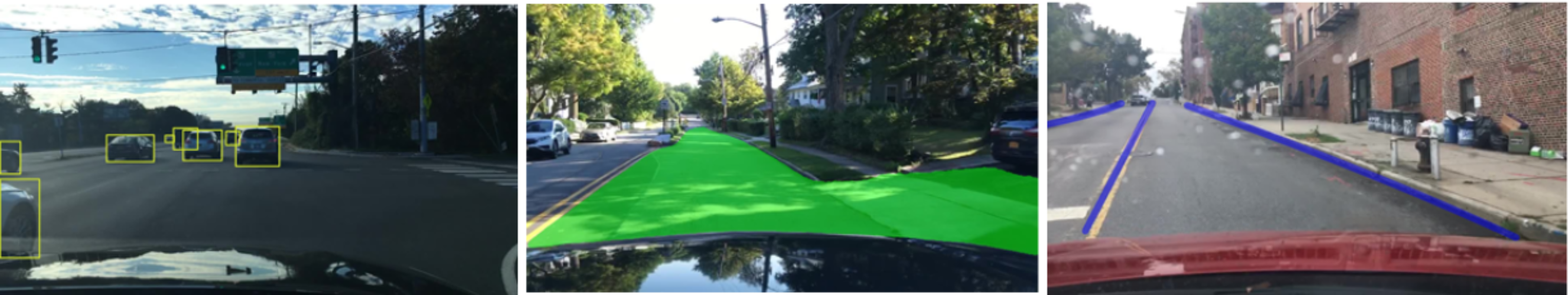}
        \caption{Examples of YOLOP \cite{wu2022yolop}}
        \label{Figure_4}
    \end{center}
\end{figure}

In this study, YOLOP is chosen as the perception module due to its efficiency, real-time performance, and integrated multi-task capability. Unlike conventional object detection algorithms that focus solely on detecting discrete objects, YOLOP simultaneously performs object detection, drivable area segmentation, and lane line detection within a unified architecture. It not only provides good lane line detection results for the AMR navigation of this study, but also provides effective assistance for subsequent roadblock detection and detection of other robots on the track. This multi-task design significantly reduces computational overhead and system complexity, making it highly suitable for embedded or resource-constrained platforms. Compared to visual SLAM algorithms, which typically require substantial computational resources and are sensitive to environmental variability, YOLOP demonstrates robust performance in outdoor scenes while maintaining real-time inference speeds. Due to hardware limitations, online visual SLAM is not feasible for this application; hence, YOLOP provides a practical and lightweight alternative that ensures adequate environmental perception for autonomous navigation.

Furthermore, although conventional image processing methods such as OpenCV-based lane detection may perform well on straight and clearly marked roads, their effectiveness significantly deteriorates in more complex scenarios. In particular, under conditions such as road curvature, dashed lane markings, and strong outdoor lighting variations, OpenCV methods are prone to instability and false detections. In contrast, YOLOP’s deep learning-based approach exhibits greater robustness and generalization in these challenging environments, making it a more reliable solution for lane detection in real-world autonomous driving scenarios.

\section{Polynomial curve fitting}
Polynomial interpolation is a fundamental method in path planning, widely used in robotics, autonomous vehicles, and computer-aided manufacturing. This approach involves constructing a polynomial function that passes through a given set of waypoints, ensuring smooth and continuous motion between them. Figure 2.3 illustrates the use of polynomial interpolation in curve fitting. This figure shows low-order polynomial interpolation. It can be seen from the figure that the fitting curve is smooth. Corresponding to autonomous navigation, the generated path points will be smooth. The primary advantage of polynomial interpolation lies in its ability to provide closed-form solutions with continuous derivatives, which are crucial for generating feasible and dynamically consistent paths. Depending on the application requirements, various orders of polynomials may be employed—such as cubic or quintic—to satisfy constraints on position, velocity, and acceleration at specific time instances.

\begin{figure}[h]
    \begin{center}
        \includegraphics[width=0.75\textwidth]{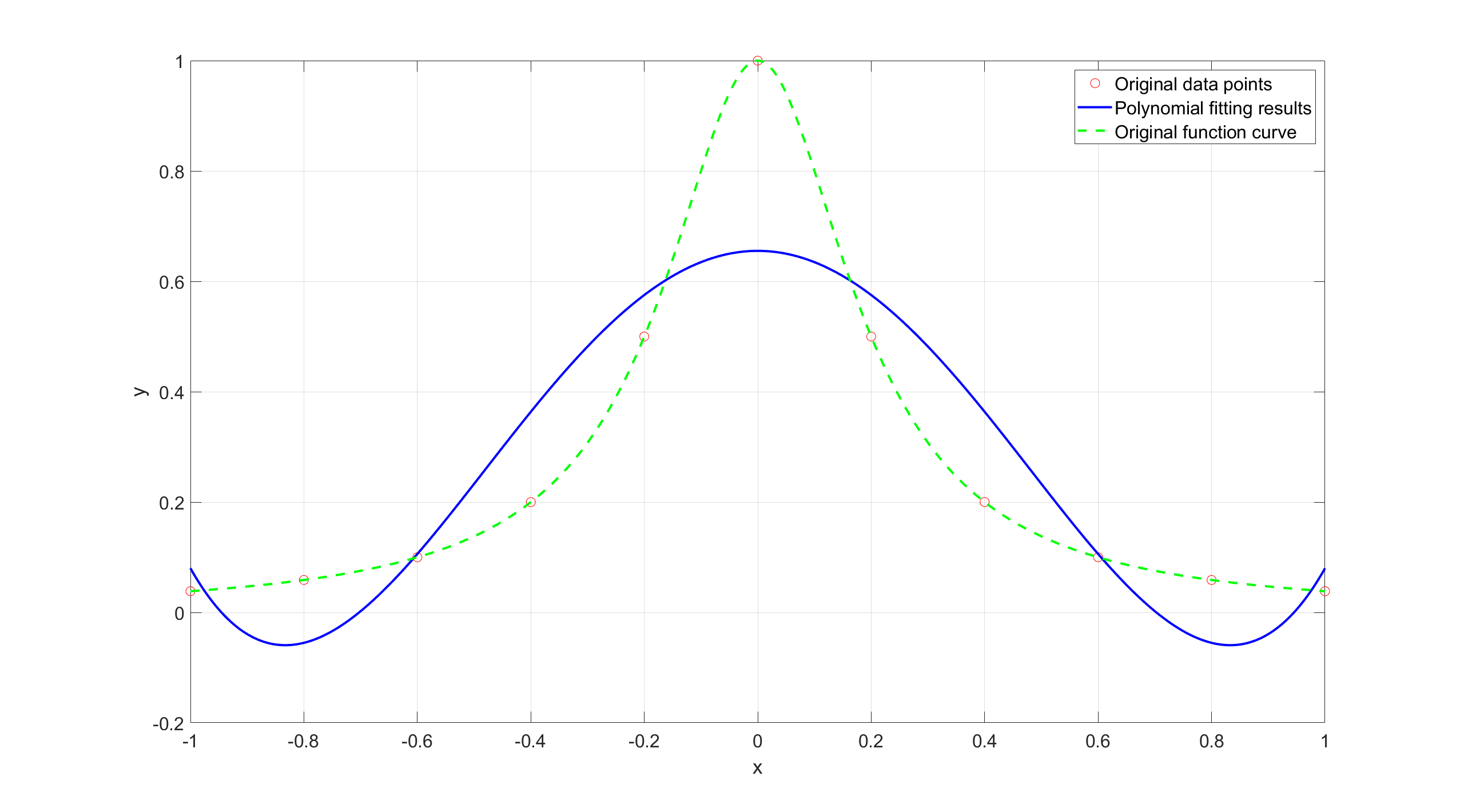}
        \caption{Polynomial interpolation in curve fitting}
        \label{Figure_4}
    \end{center}
\end{figure}

In practice, low-order polynomials like cubic polynomials are often used for point-to-point motion with velocity constraints, while higher-order polynomials, such as quintic or even septic, are preferred when smoothness up to acceleration or jerk is required. For example, quintic polynomial interpolation ensures continuous acceleration and is commonly used in manipulator path planning where minimizing abrupt dynamic changes is important \cite{sciavicco2012modelling}. The coefficients of these polynomials are typically determined by solving a system of linear equations derived from the boundary conditions, including position, velocity, and sometimes acceleration at the start and end points.

When performing polynomial interpolation on the path, the robot's posture, velocity, acceleration and other differential terms at the initial and final positions need to be given as constraints. If it is required to pass through two path points \( \theta(0)\) and \( \theta(t_0)\) within time \( t_0 \), and then consider the velocity constraints \( \dot{\theta}(0)\) and \( \dot{\theta}(t_0)\) of the initial and final points, a total of four degrees of freedom, the cubic polynomial can be completely determined
\begin{equation}
\theta(t) = a_0 + a_1 t + a_2 t^2 + a_3 t^3
\end{equation}
Polynomial coefficients are determined by the position and velocity constraints.
\begin{equation}
\begin{bmatrix}
\theta(0) \\
\theta(t_0) \\
\dot{\theta}(0) \\
\dot{\theta}(t_0)
\end{bmatrix}
=
\begin{bmatrix}
1 & 0 & 0 & 0 \\
1 & t_0 & t_0^2 & t_0^3 \\
0 & 1 & 0 & 0 \\
0 & 1 & 2t_0 & 3t_0^2
\end{bmatrix}
\begin{bmatrix}
a_0 \\
a_1 \\
a_2 \\
a_3
\end{bmatrix}
\end{equation}
Finally, we can obtain the coefficients of the cubic polynomial as
\begin{equation}
\begin{bmatrix}
a_0 \\
a_1 \\
a_2 \\
a_3
\end{bmatrix}
=
\begin{bmatrix}
\theta(0) \\
\dot{\theta}(0) \\
\frac{3}{t_0^2} \left[ \theta(t_0) - \theta(0) \right] - \frac{2}{t_0} \dot{\theta}(0) - \frac{1}{t_0} \dot{\theta}(t_0) \\
-\frac{2}{t_0^3} \left[ \theta(t_0) - \theta(0) \right] + \frac{1}{t_0^2} \left[ \dot{\theta}(0) - \dot{\theta}(t_0) \right]
\end{bmatrix}
\end{equation}
If more time differential constraints are considered, a polynomial curve of higher order is required, but the increase of the polynomial order will affect the calculation efficiency and thus reduce the real-time performance. Moreover, when using equidistant polynomial interpolation, as the order of the polynomial increases, the function analytical interval may be too small, resulting in the Runge phenomenon as shown in Figure 2.4: violent oscillations occur at both ends of the interpolation. High-order polynomial interpolation, as demonstrated by the Runge phenomenon, leads to oscillations at the endpoints of the generated path. This results in uneven path points and reduced effectiveness in autonomous navigation.

\begin{figure}[h]
    \begin{center}
        \includegraphics[width=0.75\textwidth]{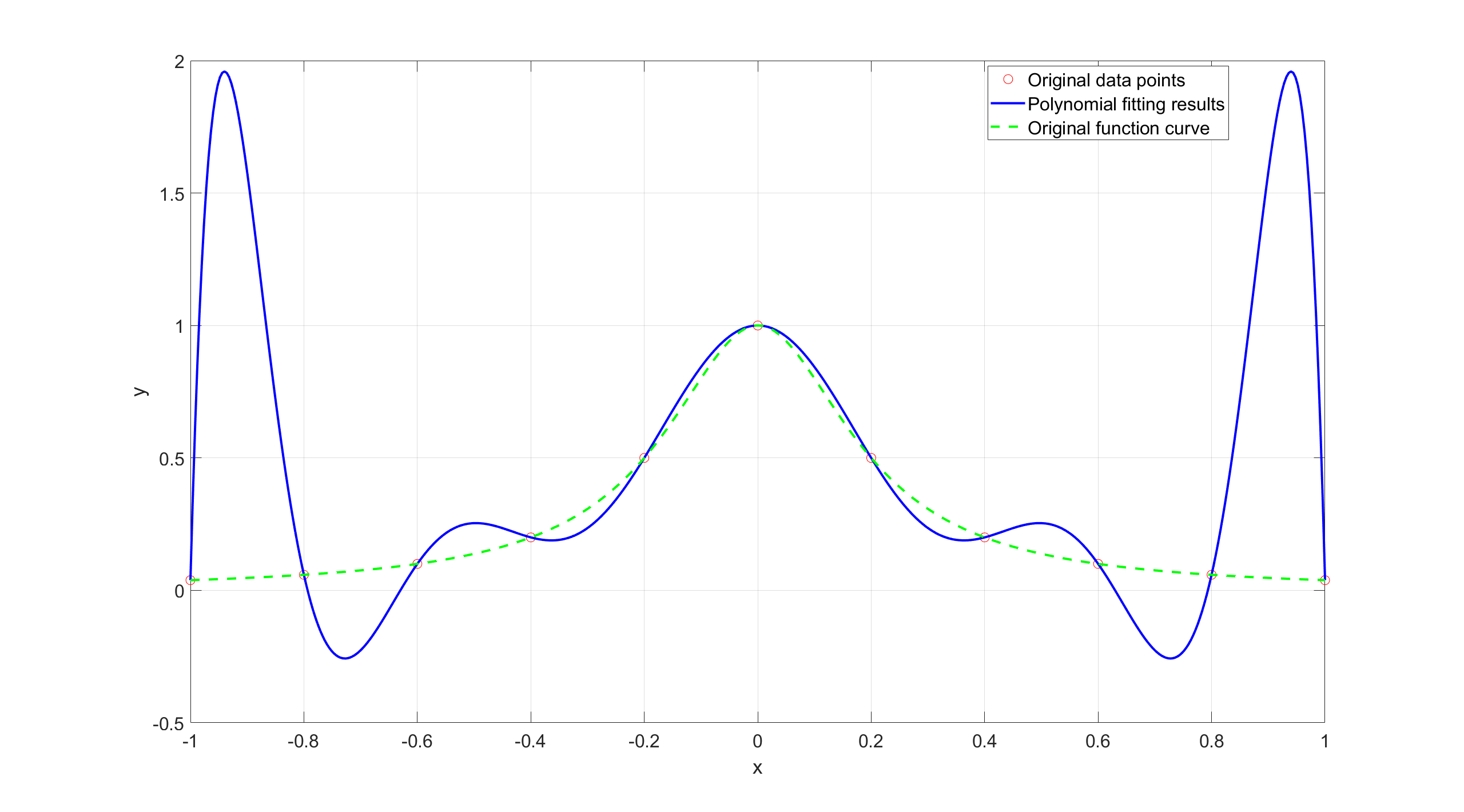}
        \caption{Runge phenomenon}
        \label{Figure_4}
    \end{center}
\end{figure}

In this study, cubic polynomial interpolation is adopted for path planning due to its balance between computational efficiency and trajectory smoothness. Compared to linear interpolation, which only ensures position continuity and often leads to abrupt changes in velocity and acceleration, cubic interpolation guarantees continuity of both position and velocity, resulting in smoother and more dynamically feasible paths for robots. On the other hand, higher-order interpolation methods, while capable of satisfying more constraints, are more susceptible to numerical instability and Runge’s phenomenon especially when fitting across multiple waypoints. Therefore, cubic interpolation offers a practical compromise, providing sufficient smoothness for path tracking while avoiding the complexity and instability associated with high-order polynomials.

\section{Lyapunov stability}
Lyapunov stability is a cornerstone concept in the analysis of dynamical systems, particularly in control theory and differential equations. Introduced by the Russian mathematician Aleksandr M. Lyapunov in the late 19th century, it provides a rigorous framework to assess whether the trajectories of a system remain bounded and close to an equilibrium point in response to small perturbations. 

Formally, an equilibrium point \( x_e \) of a dynamical system \( \dot{x} = f(x) \) is said to be Lyapunov stable if, for every \( \epsilon > 0 \), there exists a \( \delta > 0 \) such that \( \|x(0) - x_e\| < \delta \) implies \( \|x(t) - x_e\| < \epsilon \) for all \( t \geq 0 \). If, in addition, \( \|x(t) - x_e\| \to 0 \) as \( t \to \infty \), the equilibrium is said to be \textit{asymptotically stable}. These definitions do not require explicit solutions of the system, making Lyapunov’s approach broadly applicable to nonlinear and time-invariant systems \cite{khalil2002nonlinear}.

The first method proposed by Lyapunov, often referred to as the indirect method, assesses the stability of a nonlinear system by examining its linear approximation near an equilibrium point. Given a nonlinear autonomous system
\begin{equation}
\dot{x} = f(x), \quad f(0) = 0,
\end{equation}
the Jacobian matrix \( A = \left. \frac{\partial f}{\partial x} \right|_{x=0} \) is computed at the equilibrium. The local behavior of the system is then approximated by the linear system
\begin{equation}
\dot{x} = Ax.
\end{equation}
Lyapunov's first method states that if all eigenvalues of \( A \) have negative real parts, that is, \( \text{Re}(\lambda_i) < 0 \) for all \( i \), then the equilibrium at the origin is asymptotically stable. If any eigenvalue has a positive real part, the equilibrium is unstable. However, if any eigenvalue lies on the imaginary axis, the method is inconclusive. This method is particularly useful due to its simplicity and the availability of linear algebraic tools, although it only provides local stability information and is limited to systems with differentiable right-hand sides \cite{slotine1991applied}.

The second, or direct method, is more powerful and broadly applicable, especially for nonlinear systems. It avoids linearization by constructing a scalar Lyapunov function \( V(x): \mathbb{R}^n \to \mathbb{R} \), analogous to an energy function, satisfying the following properties:
\begin{itemize}
  \item \( V(0) = 0 \), and \( V(x) > 0 \) for all \( x \neq 0 \) (positive definite);
  \item The time derivative \( \dot{V}(x) = \frac{\partial V}{\partial x} f(x) \) is non-positive, i.e., \( \dot{V}(x) \leq 0 \) (negative semi-definite).
\end{itemize}
If such a function exists, then the equilibrium at the origin is Lyapunov stable. If \( \dot{V}(x) < 0 \) (negative definite), the equilibrium is asymptotically stable. In practice, candidate Lyapunov functions are often constructed using quadratic forms, such as
\begin{equation}
V(x) = x^\top P x,
\end{equation}
where \( P = P^\top > 0 \) is a positive definite matrix. The choice of \( P \) can be determined using Lyapunov’s matrix equation for linear systems or through heuristic design in nonlinear contexts. The second method is particularly powerful for analyzing global stability and is extensively employed in modern nonlinear control system design, including adaptive and robust control frameworks \cite{vidyasagar2002nonlinear}.

Lyapunov stability theory offers a fundamental framework for assessing the behavior of dynamical systems near equilibrium. While the first method provides local stability analysis through system linearization, it is limited in handling nonlinear dynamics. In contrast, Lyapunov’s second method allows direct stability analysis without linearization and is better suited for nonlinear systems. Therefore, this study adopts the second method to design the path-tracking controller, as it enables the construction of control laws that ensure asymptotic stability and robustness against model uncertainties and external disturbances, which are critical for reliable autonomous navigation. Compared to traditional methods such as PID control or pure pursuit algorithms which often rely on heuristic tuning and may struggle with stability and performance under complex dynamic conditions. The Lyapunov-based approach offers a more systematic and theoretically grounded framework for ensuring stability and convergence, particularly in nonlinear and time-varying environments.
\chapter{Modeling}

\section{Robot kinematic modeling}
The nomenclature of the main variables in this section is shown in Table 3.1.
\begin{table}[ht]
\centering
\caption{Description of variables and their units}
\begin{tabular}{|c|l|c|}
\hline
Variable & Description & Unit \\
\hline
$x$ & Robot geometric center horizontal coordinate & m \\
$y$ & Robot geometric center Longitudinal coordinate & m \\
$\phi$ & Robot heading angle & rad \\
$r$ & Wheel radius & m \\
$r_m$ & Robot movement radius & m \\
$d$ & Wheelbase & m \\
$v$ & Robot linear velocity & m/s \\
$\omega$ & Robot's angular velocity & rad/s \\
$\omega_{L}$ & Left wheel angular velocity & rad/s \\
$\omega_{R}$ & Right wheel angular velocity & rad/s \\
$v_{L}$ & Left wheel linear velocity & m/s \\
$v_{R}$ & Right wheel linear velocity & m/s \\
$t$ & Time index & s \\
\hline
\end{tabular}
\end{table}

This section introduces the kinematic model of the AI Formula. As a three-wheel differential model, it is important to note that the front wheels are responsible for both propulsion and steering, while the rear wheels remain passive. Therefore, the robot model is simplified to a two-wheel differential model to facilitate the subsequent controller design. The robot under consideration follows a structural configuration to a two-wheel differential drive, as illustrated in Figure 3.1. In this model, point \(O\) represents the geometric center of the simplified model. The position of the geometric center is given by the coordinates \(O (x, y)\), where \(v\) and \(\omega\) represent the robot’s linear and angular velocities, respectively, and \(\phi\) denotes the heading angle.

\begin{figure}[h]
    \begin{center}
        \includegraphics[width=0.75\textwidth]{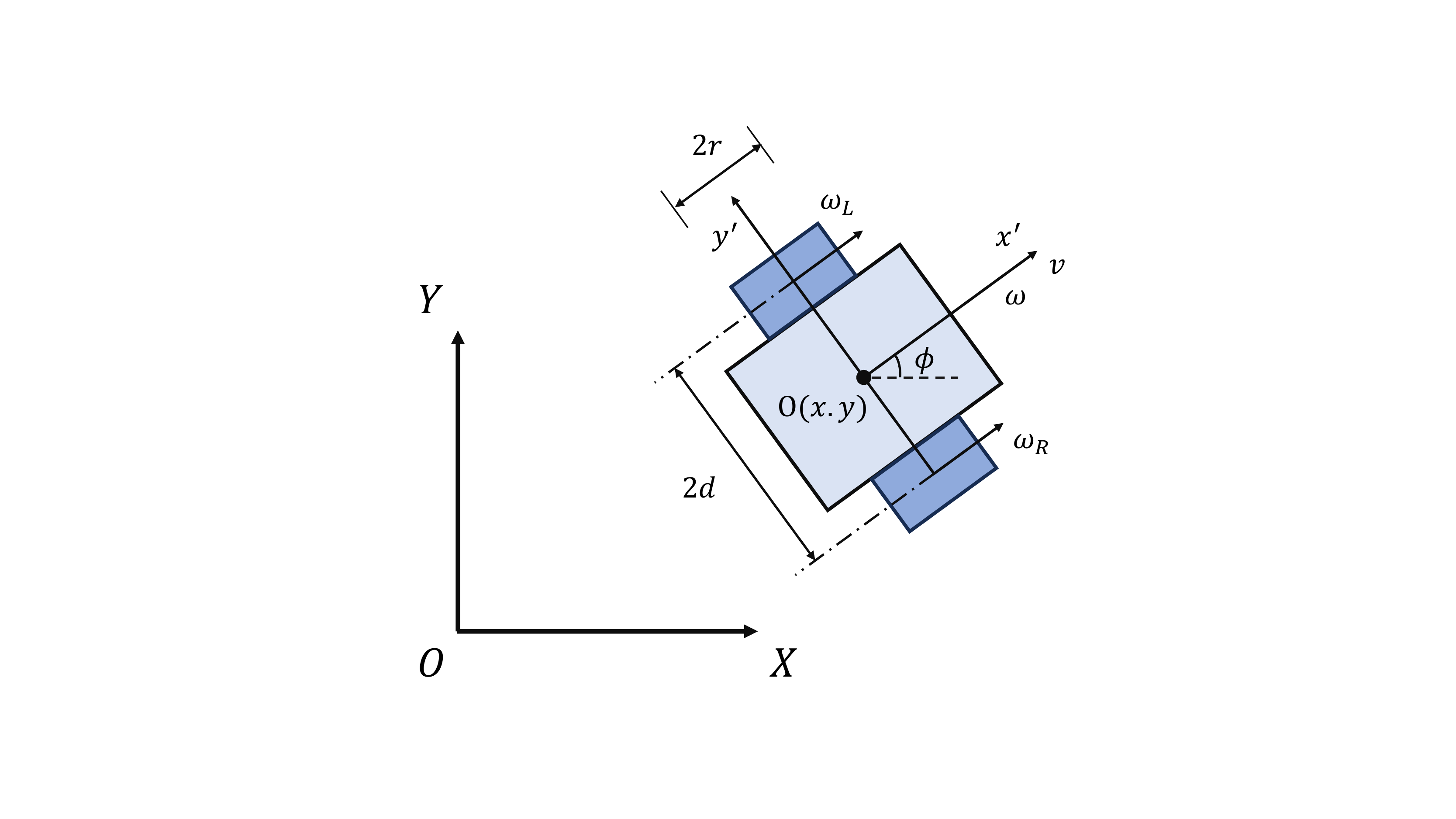}
        \caption{Two-wheel differential model}
        \label{Figure_5}
    \end{center}
\end{figure}

In this model, the robot's geometric center and mass center are assumed to coincide, and the coincidence point is at the midpoint of the wheel axle. Let \(p=\begin{bmatrix}x&y&\phi\end{bmatrix}^T\) represent the robot's posture in the global coordinate system \(xOy\), and \(M = (x, y)\) represents the coordinates of the robot's mass center in the global coordinate system. Let the robot's wheel diameter be \(2r\), and the distance between the two driving wheels be \(2d\).

Let the local coordinates of the robot be \(x^{\prime}Oy^{\prime}\), then define the robot's linear velocity \(v\) as its velocity along the \(x^{\prime}\)-axis, the heading angle \(\phi\) as the angle between the \(x^{\prime}\)-axis and the \(x\)-axis of the global coordinate system, the angular velocity \(\omega\) as the instantaneous angular rate of rotation of the \(x^{\prime}\)-axis around point \(O\), and the entire system uses \(u=\begin{bmatrix}v&\omega\end{bmatrix}^T\) as the input control variable. Let \(\omega_{L}\), \(\omega_{R}\) and \(v_{L}\), \(v_{R}\) be the angular velocity, linear velocity of the left and right driving wheels of the robot, respectively, and also the drive variable.

Below, we'll analyze the relationship between the control variable and the drive variable. First, we have
\begin{equation}
v = \frac{v_L + v_R}{2}
\end{equation}
As shown in the Figure 3.2, within time \(\Delta t \rightarrow 0\), the robot moves from posture \(p_1\) to posture \(p_2\), and the heading angle \(\phi\) changes by \(\Delta\phi\). The right wheel moves \(\Delta l = (v_R - v_L) \Delta t\) more than the left wheel. Then \(\Delta \phi \approx \sin \Delta \phi = \frac{\Delta l}{2d}\), thus obtaining
\begin{equation}
\omega = \frac{\Delta \phi}{\Delta t} = \frac{v_R - v_L}{2d}
\end{equation}
This allows us to determine the robot's motion radius
\begin{equation}
r_m = \frac{v}{\omega} = \frac{v_R + v_L}{v_R - v_L} d
\end{equation}

\begin{figure}[h]
    \begin{center}
        \includegraphics[width=0.75\textwidth]{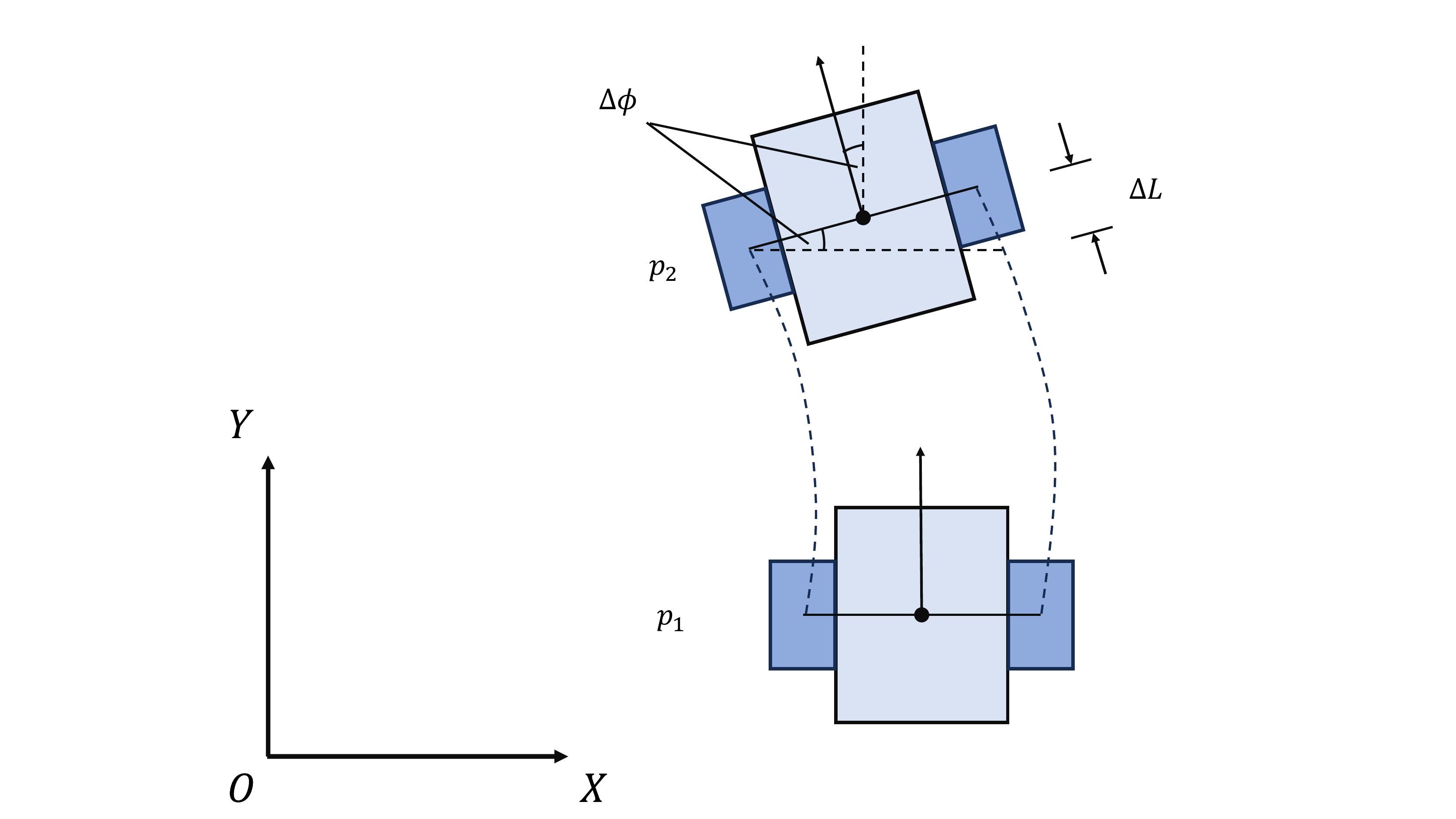}
        \caption{Kinematic model of two-wheel differential robot}
        \label{Figure_6}
    \end{center}
\end{figure}

In the end, the relationship between the robot's control and drive variables can be represented by a matrix as
\begin{equation}
\begin{bmatrix}
v \\
\omega
\end{bmatrix}
=
\begin{bmatrix}
\frac{1}{2} & \frac{1}{2} \\
\frac{1}{2d} & -\frac{1}{2d}
\end{bmatrix}
\begin{bmatrix}
v_R \\
v_L
\end{bmatrix}
\end{equation}

The kinematic model of the two-wheel differential drive relates the control inputs to the time derivatives of the vehicle’s position and orientation. The center of rotation for the angular velocity is located at the geometric center. The following relationship is derived from the kinematic equations of the two-wheel differential model:
\begin{equation}
\dot{\mathbf{p}} =
\begin{bmatrix}
\dot{x} \\
\dot{y} \\
\dot{\phi}
\end{bmatrix}
=
\begin{bmatrix}
\cos \phi & 0 \\
\sin \phi & 0 \\
0 & 1
\end{bmatrix}
\begin{bmatrix}
v \\
\omega
\end{bmatrix}
\end{equation}
It can also be further written as Eq. (3.6) based on the relationship between the control variable and the drive variable.
\begin{equation}
\dot{\mathbf{p}} =
\begin{bmatrix}
\dot{x} \\
\dot{y} \\
\dot{\theta}
\end{bmatrix}
=
\begin{bmatrix}
\cos \phi & 0 \\
\sin \phi & 0 \\
0 & 1
\end{bmatrix}
\begin{bmatrix}
\frac{1}{2} & \frac{1}{2} \\
\frac{1}{2d} & -\frac{1}{2d}
\end{bmatrix}
\begin{bmatrix}
v_R \\
v_L
\end{bmatrix}
\end{equation}
The above equation shows that the autonomous mobile robot is a nonholonomic system with a 3-dimensional state vector and 2 control inputs. Therefore, the autonomous mobile robot has motion constraints that cannot move omnidirectionally, and path smoothness and continuity need to be coordinated in path planning. This is also the reason why we introduce the path tracking algorithm based on the Lyapunov controller.

\section{Geometric path kinematic modeling}
The nomenclature of the main variables in this section is shown in Table 3.2.
\begin{table}[ht]
\centering
\caption{Description of variables and their units}
\begin{tabular}{|c|l|c|}
\hline
Variable & Description & Unit \\
\hline
$x_t$ & Target point horizontal coordinate & m \\
$y_t$ & Target point Longitudinal coordinate & m \\
$\phi_t$ & Target point heading angle & rad \\
$v_t$ & Robot target velocity & m/s \\
$\alpha$ & Defined angular error & rad \\
$\beta$ & Defined angular error & rad \\
\hline
\end{tabular}
\end{table}

The geometric path to be studied is illustrated in the accompanying Figure 3.3. To facilitate the system description, the following notations are adopted. Global coordinate frame is denoted as \(xOy\), which defines the workspace. The robot's geometric center is represented by the point \(O(x, y)\), whose position evolves over time. The robot moves with a linear velocity \(v\) and an angular velocity \(\omega\), and its motion direction is given by the heading angle \(\phi\). The target point to be tracked is denoted as \(O_t(x_t, y_t)\), with a corresponding target velocity \(v_t\) and target heading direction \(\phi_t\). The scalar \(\rho\) denotes the Euclidean distance between the robot and the target point, while \(\theta\) represents the angle formed by the vector from the robot's center \(O\) to the target position \(O_t\) relative to the global frame. These notations form the basis for describing the relative motion and formulating the control laws.

\begin{figure}[h]
    \begin{center}
        \includegraphics[width=0.75\textwidth]{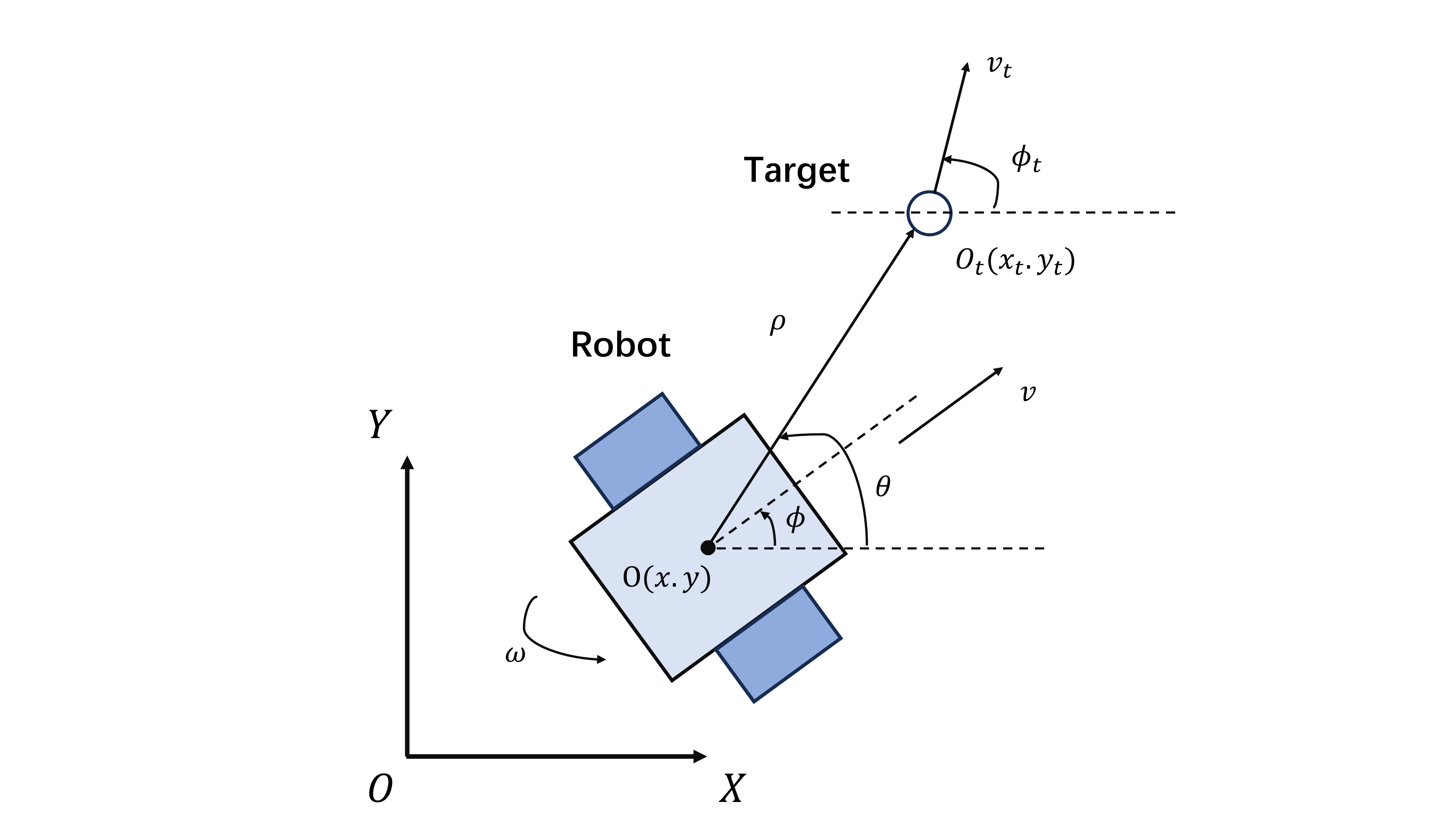}
        \caption{Mobile robot tracking a moving target point}
        \label{Figure_7}
    \end{center}
\end{figure}

The robot's kinematics is described by the following set of differential equations, which characterize the motion of a non-holonomic mobile robot based on its velocity inputs: 
\begin{align}
\dot{x} &= v \cos \phi \\
\dot{y} &= v \sin \phi \\
\dot{\phi} &= \omega
\end{align}

The relative positions and orientations of the robot and the moving target point can be expressed through the following set of geometric relationships:
\begin{align}
\rho \cos \theta &= x_t - x \\
\rho \sin \theta &= y_t - y \\
\alpha &= \theta - \phi \\
\beta &= \theta - \phi_t
\end{align}

Their derivatives with respect to time \(t\) are 
\begin{align}
\dot{\rho} &= v_t \cos \beta - v \cos \alpha \\
\dot{\alpha} &= v \frac{\sin \alpha}{\rho} - v_t \frac{\sin \beta}{\rho} - \omega, \quad \rho \ne 0 \\
\dot{\beta} &= v \frac{\sin \alpha}{\rho} - v_t \frac{\sin \beta}{\rho} - \dot{\phi}_t, \quad \rho \ne 0
\end{align}

\(\dot{\phi}_t\) is calculated from the three look ahead points shown in the figure, and its essence is the target angular velocity \(\omega_t\). By introducing the target angular velocity, the robot can drive to the next target point more smoothly when it reaches the target point.

$\dot{\phi}_t$ is computed based on three look-ahead points, as illustrated in the Figure. 3.4 . Essentially, $\dot{\phi}_t$ represents the target angular velocity, denoted as $\omega_t$, which serves as a predictive control reference for the robot’s rotational motion. By incorporating this target angular velocity into the control framework, the robot is able to adjust its heading more smoothly as it approaches each designated target point along the path. This predictive adjustment enhances the continuity and fluidity of motion, thereby improving trajectory tracking performance and overall navigation stability.

\begin{align}
\phi_{AB} &= arctan(\frac{y_B-y_A}{x_B-x_A}) \\
\phi_{BC} &= arctan(\frac{y_C-y_B}{x_C-x_B}) \\
\dot{\phi}_t &= \frac{\phi_{BC} - \phi_{AB}}{\Delta t}
\end{align}

\begin{figure}[h]
    \begin{center}
        \includegraphics[width=0.45\textwidth]{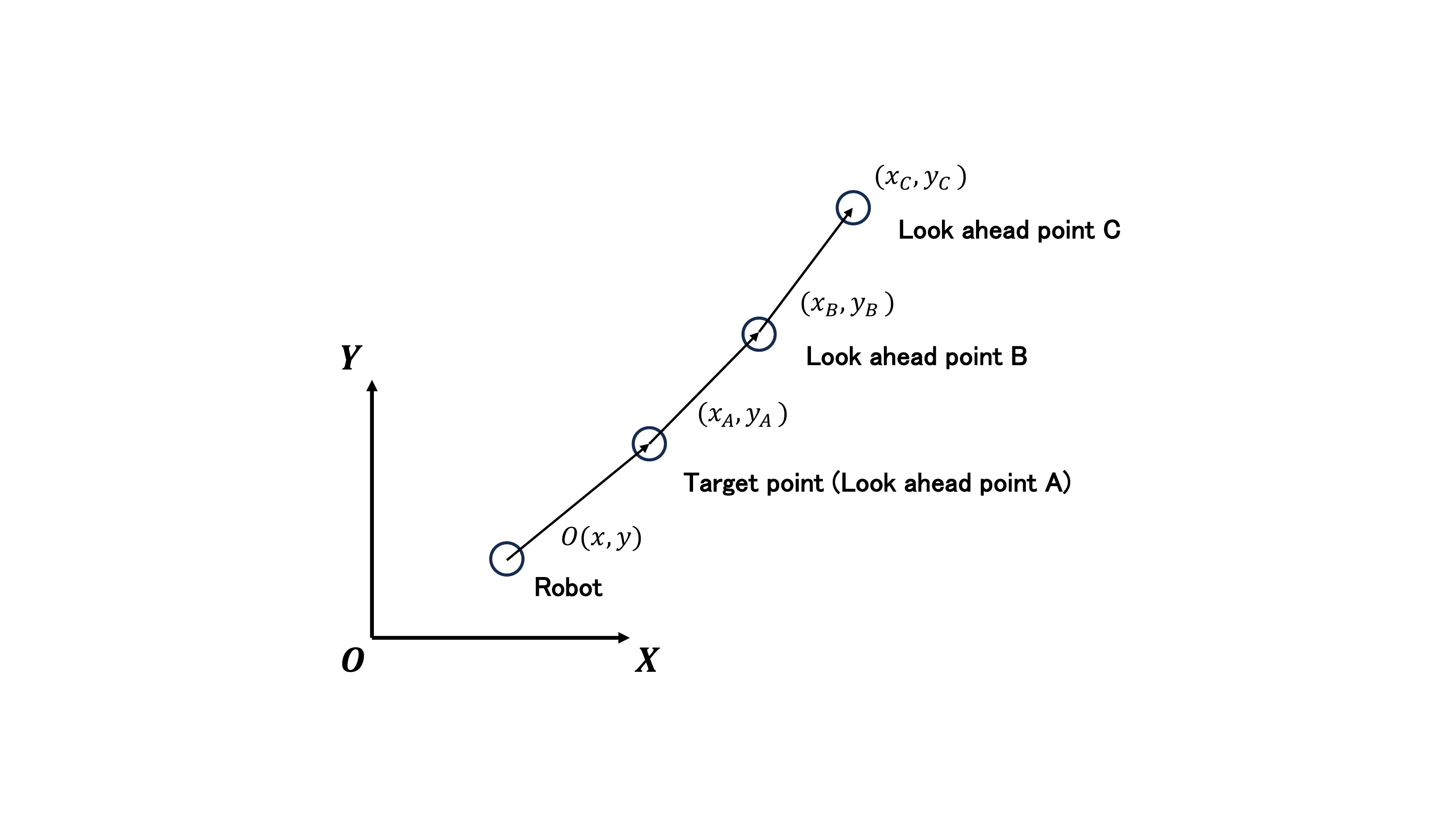}
        \caption{Three look ahead points}
        \label{Figure_7}
    \end{center}
\end{figure}

Eq. (3.14-3.16) represent the kinematic model of the geometric path in terms of a new set of state variables expressed in polar coordinates: \(\rho\), \(\alpha\), and \(\beta\). Here, \(\rho\) denotes the linear tracking error, measuring the distance between the robot and the moving target. The variables \(\alpha\) and \(\beta\) represent angular tracking errors, with \(\alpha\) indicating the misalignment between the robot’s heading and the line-of-sight to the target, and \(\beta\) capturing the angular deviation between the target’s heading and the same reference direction. These polar-coordinate variables are essential for formulating error dynamics and designing effective tracking controllers.

\chapter{Robot Autonomous Navigation Algorithms}

\section{YOLOP}
The nomenclature of the main variables in this section is shown in Table 4.1.
\begin{table}[ht]
\centering
\caption{Description of variables}
\begin{tabular}{|c|l|c|}
\hline
Variable & Description \\
\hline
$ x, y $ & Original pixel coordinates in the input image \\
$ x', y' $ & Rescaled coordinates after letterbox padding \\
$ r $ & Image resizing ratio \\
$ p_x, p_y $ & Padding values in horizontal and vertical directions \\
$ x_{\text{norm}} $ & Normalized pixel value \\
$ \mu $ & Mean for normalization (per channel) \\
$ \sigma $ & Standard deviation for normalization (per channel) \\
$ I $ & Input image tensor \\
$ F $ & Shared feature map from the encoder \\
$ O_{\text{det}} $ & Output of the object detection head \\
$ O_{\text{ll}} $ & Output of the lane line segmentation head \\
$ s_i $ & Confidence score for the $i$-th bounding box \\
$ b_i $ & $i$-th predicted bounding box \\
$ \tau_{\text{conf}} $ & Confidence threshold for filtering \\
$ \tau_{\text{IoU}} $ & IoU threshold for Non-Maximum Suppression \\
$ \text{IoU}(b_i, b_j) $ & Intersection-over-Union between boxes $b_i$ and $b_j$ \\
$ \hat{y}_{\text{ll}}(i,j) $ & Final predicted class at pixel $(i,j)$ \\
$ P_c(i,j) $ & Predicted probability of class $c$ at pixel $(i,j)$ \\
$ \mathcal{L}_{\text{CE}} $ & Cross-entropy loss \\
$ \mathcal{L}_{\text{IoU}} $ & IoU-based segmentation loss \\
$ TP, FP, FN $ & True positives, false positives, false negatives \\
\hline
\end{tabular}
\end{table}

This section outlines the implementation of a real-time multi-task perception system based on the YOLOP architecture. The system performs joint object detection and lane line segmentation on single-frame RGB inputs. 

Input images are first resized using the letterbox method, which preserves the original aspect ratio while fitting the network’s input resolution (typically 640\,$\times$\,640). Letting $r$ denote the resizing ratio and $(p_x, p_y)$ the padding applied to the width and height respectively, the transformation of spatial coordinates is given by:
\begin{equation}
(x', y') = (r x + p_x,\; r y + p_y)
\end{equation}

This operation ensures consistent spatial correspondence between input and output domains. After resizing, each pixel value is normalized per channel using dataset-specific mean $\mu$ and standard deviation $\sigma$:
\begin{equation}
x_{\text{norm}} = \frac{x / 255 - \mu}{\sigma}
\end{equation}

The resulting tensor is converted to floating-point precision  and passed to the model for inference.

The implementation employs the YOLOP architecture as proposed in the original work \cite{wu2022yolop}, which integrates a shared backbone with three task-specific heads. Given an input image $I$, the model produces a joint output:
\begin{equation}
F = \text{Backbone}(I), \quad (O_{\text{det}}, O_{\text{da}}, O_{\text{ll}}) = \text{Heads}(F)
\end{equation}
where $F$ denotes the shared feature representation extracted by the CSPDarknet backbone. The detection head $O_{\text{det}}$ predicts bounding boxes and class probabilities, while $O_{\text{ll}}$ provides a semantic segmentation map for lane markings. In the current implementation, the drivable area head $O_{\text{da}}$ is omitted.

The detection head outputs anchor-based predictions over multiple spatial scales. Postprocessing involves confidence filtering and Non-Maximum Suppression (NMS). Given a confidence threshold $\tau_{\text{conf}}$, candidate detections $b_i$ with score $s_i$ are retained if:
\begin{equation}
s_i \geq \tau_{\text{conf}}
\end{equation}
Redundant detections are suppressed using the standard IoU metric:
\begin{equation}
\text{IoU}(b_i, b_j) = \frac{|b_i \cap b_j|}{|b_i \cup b_j|}
\end{equation}
Boxes with IoU exceeding a predefined threshold $\tau_{\text{IoU}}$ are eliminated in favor of the higher-scoring detection. The remaining bounding box coordinates are rescaled to match the original image resolution via the inverse of the letterbox transformation:
\begin{equation}
(x, y) = \frac{x' - p_x}{r}
\end{equation}

The lane line segmentation head produces a coarse pixel-wise prediction map of shape $[1, C=2, H/8, W/8]$, where $C$ denotes the number of semantic classes (lane vs. background). The predicted feature map is first cropped to remove padding, then upsampled to the original image resolution using bilinear interpolation. Final segmentation labels are obtained by selecting the most probable class per pixel:
\begin{equation}
\hat{y}_{\text{ll}}(i,j) = \arg\max_{c \in \{0,1\}} P_c(i,j)
\end{equation}

This operation yields a binary segmentation mask indicating lane line presence. During training, this head is optimized using a composite loss combining cross-entropy and IoU-based terms:
\begin{equation}
\mathcal{L}_{\text{ll-seg}} = \mathcal{L}_{\text{CE}} + \mathcal{L}_{\text{IoU}}, \quad
\mathcal{L}_{\text{IoU}} = 1 - \frac{TP}{TP + FP + FN}
\end{equation}

Although the present implementation focuses on inference, the underlying architecture and behavior remain consistent with the supervised multi-task learning objective defined in the original model.

This implementation preserves the computational structure of the YOLOP model while enabling efficient real-time inference. Each processing stage is tightly coupled with the mathematical formulation presented in the original work, ensuring both architectural fidelity and practical performance.

\section{Polynomial curve fitting}
The nomenclature of the main variables in this section is shown in Table 4.2.
\begin{table}[ht]
\centering
\caption{Description of variables}
\begin{tabular}{|c|l|c|}
\hline
Variable & Description \\
\hline
\( (u, v) \) & Pixel coordinates in image frame. \\
\( K \) & Camera intrinsic matrix. \\
\( p_{\text{cam}} \) & Point in camera coordinate frame\\
\( p_{\text{veh}} \) & Point in vehicle coordinate frame\\
\( \lambda \) & Depth scale factor from projection\\
\( P_i \) & Original detected 3D point along lane polyline\\
\( S_i \) & Cumulative arc length up to the \(i\)-th point\\
\( \Delta s \) & Resampling interval in arc length\\
\( L_k \) & Target arc length for the \(k\)-th resampled point\\
\( Q_k \) &  Interpolated 3D point after resampling\\
\( (x_i, y_i) \) & 2D projection of resampled lane points for fitting\\
\( a \) & Coefficient vector of the cubic polynomial\\
\( V \) & Vandermonde matrix constructed from \(x_i\)\\
\( y \) & Vector of corresponding \(y_i\) values\\
\( Q, R \) & Orthogonal and upper triangular matrices from QR decomposition\\
\hline
\end{tabular}
\end{table}

This section presents the mathematical foundation of the lane line reconstruction algorithm. The method consists of two core components: arc-length-based resampling of parameterized polylines and cubic polynomial curve fitting. These steps transform noisy and unevenly distributed raw lane detection points into a smooth and uniformly sampled trajectory suitable for downstream path planning and control.

The process begins by projecting 2D pixel coordinates from the image space into the 3D vehicle coordinate system. Given a pixel location \((u, v)\), a camera intrinsic matrix \(K\), and an extrinsic transformation matrix \(T_{\text{veh} \leftarrow \text{cam}}\), the 3D point \(p_{\text{veh}} \in \mathbb{R}^3\) is obtained via:

\begin{equation}
K^{-1}
\begin{bmatrix} u \\ v \\ 1 \end{bmatrix}
= \lambda
\begin{bmatrix} X_c \\ Y_c \\ Z_c \end{bmatrix}
\Rightarrow
p_{\text{cam}} = (X_c, Y_c, Z_c)^\top,
\end{equation}
where \(\lambda\) is a depth scaling factor determined via a flat ground assumption (e.g., \(Z_c = 0\)) or by depth sensors.

\begin{equation}
p_{\text{veh}} = T_{\text{veh} \leftarrow \text{cam}} \cdot 
\begin{bmatrix} p_{\text{cam}} \\ 1 \end{bmatrix}.
\end{equation}

To suppress outliers and improve fitting quality, points outside the region of interest (ROI) \([x_{\min}, x_{\max}] \times [y_{\min}, y_{\max}]\) are excluded.

Detected lane points often appear as sparse or irregularly spaced polylines in the vehicle frame. To enhance stability, these polylines are resampled at constant arc-length intervals. Let the original 3D point sequence be \(\{P_i\}_{i=0}^{n-1}\). The cumulative arc-length \(S_i\) is computed iteratively:

\begin{equation}
S_0 = 0, \quad S_i = S_{i-1} + \|P_i - P_{i-1}\|_2.
\end{equation}

Define the uniform resampling arc-length targets:
\begin{equation}
L_k = k \cdot \Delta s, \quad 
k = 0, 1, \ldots, 
\left\lfloor \frac{S_{n-1}}{\Delta s} \right\rfloor.
\end{equation}

For each \(L_k\), locate segment \([S_j, S_{j+1}]\) and compute interpolated point \(Q_k\):
\begin{equation}
Q_k = (1 - t){P_j + t P}_{j+1}, \quad
t = \frac{L_k - S_j}{S_{j+1} - S_j}.
\end{equation}

This procedure produces arc-length-equidistant points, ensuring consistent geometric distribution for subsequent fitting.

To construct a continuous lane curve, we fit a third-order polynomial to the resampled set \(\{(x_i, y_i)\}\):
\begin{equation}
p(x) = a_0 + a_1 x + a_2 x^2 + a_3 x^3,
\end{equation}
where \(\mathbf{a} = [a_0, a_1, a_2, a_3]^\top\) are the coefficients to be estimated. \( p(x) \) denote the cubic polynomial function obtained through curve fitting, where \( x_i \) and \( y_i \) represent the horizontal and vertical coordinates of the input sample points, respectively. The value \( p(x_i) \) corresponds to the predicted output of the fitting function at point \( x_i \). The closer \( p(x_i) \) is to \( y_i \), the smaller the fitting error, indicating a more accurate and effective fitting performance.

The least-squares problem is formulated by constructing the Vandermonde matrix \(V \in \mathbb{R}^{n \times 4}\) and observation vector \(\mathbf{y}\):
\begin{equation}
V = 
\begin{bmatrix}
1 & x_1 & x_1^2 & x_1^3 \\
1 & x_2 & x_2^2 & x_2^3 \\
\vdots & \vdots & \vdots & \vdots \\
1 & x_n & x_n^2 & x_n^3
\end{bmatrix}, \quad
\mathbf{y} = 
\begin{bmatrix}
y_1 \\ y_2 \\ \vdots \\ y_n
\end{bmatrix}.
\end{equation}

The fitting process involves minimizing the squared error between the predicted values \(p(x_i)\) and the observed values \(y_i\)

\begin{equation}
\min_{\mathbf{a}} \sum_{i=1}^{N} \left( p(x_i) - y_i \right)^2
\end{equation}

To solve this robustly, QR decomposition with column pivoting is applied:
\begin{equation}
V = Q R \Rightarrow \mathbf{a} = R^{-1} Q^\top \mathbf{y},
\end{equation}
thus avoiding explicit inversion of \(V^\top V\).

Finally, the polynomial is evaluated over a uniformly spaced set of \(x\)-coordinates:
\begin{equation}
y_i = a_0 + a_1 x_i + a_2 x_i^2 + a_3 x_i^3.
\end{equation}

The result will be smooth and regularly spaced lane coordinates\((x_i,y_i) \). suitable for integration into motion planning systems. Then, based on the generated left and right lane line path points, the sum of their horizontal coordinates is divided by 2 to obtain the center line of the lane, and finally we can obtain the required target point.

Due to the many different situations in the actual course, the robot may fail to detect two lane lines in the image. Therefore, we supplement the algorithm so that the robot can navigate autonomously even when the lane lines are insufficient or missing. The handling logic is as follows:

\begin{enumerate}
    \item When both lane lines are detected, navigate according to the generated target point.
    \item When one lane line is missing, generate the lane line on the missing side based on the known lane width ($3.5m$), then calculate the center line to obtain the target point.
    \item When no lane line is detected in the image, the robot drives forward at the preset minimum speed $v_\text{min}$ (which will be introduced in Chapter 5).
\end{enumerate}

This section has presented a mathematically grounded pipeline for lane reconstruction, involving coordinate transformation, arc-length resampling, and cubic polynomial fitting. The use of QR-based least-squares optimization enhances robustness to outliers and ill-conditioned data, while arc-length parametrization improves the geometric regularity of the input. Together, these methods enable the conversion of sparse detections into a high-quality, continuous lane model.

\section{Lyapunov-based control}
The nomenclature of the main scalars and preset parameters in this section is shown in Table 4.3.
\begin{table}[ht]
\centering
\caption{Description of scalars and preset parameters}
\begin{tabular}{|c|l|c|}
\hline
\textbf{Variable} & \textbf{Description} \\
\hline
$V$ & The total Lyapunov function \\
$V_1$ & Lyapunov function term related to linear tracking error \\
$V_2$ & Lyapunov function term related to angular tracking error \\
$k_1$ & Weight on $\alpha$ in Lyapunov $V_2$ \\
$k_2$ & Weight on $\beta$ in Lyapunov $V_2$r \\
$\lambda_v$ & Gain that shrinks distance $\rho$ \\
$\lambda_a$ & Gain that drives heading angle error to 0\ \\
\hline
\end{tabular}
\end{table}

The primary objective of the controller is to enable the robot to accurately track a target point that changes its position dynamically over time. To achieve this goal, a controller grounded in Lyapunov stability theory is proposed. This controller is designed to compute both the linear and angular velocities required for the robot to adjust its motion in real-time. By continuously regulating the robot’s position and orientation, the controller ensures that the robot follows the real-time changing path target points effectively and stably. The design and implementation of the controller are firmly based on the kinematic model of the robot, which provides a mathematical representation of its motion characteristics. Through the utilization of this model, the controller can precisely determine the necessary control inputs to guide the robot in a manner that guarantees convergence to the desired path.

The following candidate Lyapunov function is chosen for controller design to facilitate the stability analysis and guide the formulation of the control laws:
\begin{align}
V &= V_1 + V_2 \\
V_1 &= \frac{1}{2} \rho^2 \\
V_2 &= \frac{1 - \cos \alpha}{k_1} + \frac{1 - \cos \beta}{k_2}
\end{align}
where \(V_1\) and \(V_2\) are scalar functions specifically designed to quantify the linear and angular tracking errors, respectively. The term \(V_1\) is associated with the translational deviation of the robot from the desired target position, while \(V_2\) reflects the orientation error between the robot’s current heading and the desired direction. By appropriately constructing \(V_1\) and \(V_2\), the Lyapunov function \(V\) serves as a measure of the overall tracking performance. Ensuring that the time derivative of \(V\) is negative definite guarantees the asymptotic convergence of both position and orientation errors, thereby validating the stability and effectiveness of the proposed controller.

Differentiating \(V_1\) with respect to time \(t\), and taking into account the system dynamics as described in Eq. (3.14), we obtain the time derivative \(\dot{V}_1\),
\begin{equation}
\dot{V}_1 = \rho (v_t \cos \beta - v \cos \alpha)
\end{equation}

To ensure the non-positivity of \(\dot{V}_1\), the linear velocity \(v\) is designed in such a way that it counteracts the growth of the linear tracking error. The control law for \(v\) is therefore derived as follows:
\begin{equation}
v = \frac{v_t \cos \beta}{\cos \alpha} + \lambda_v \rho \cos \alpha
\end{equation}
where \(\lambda_v > 0\) is a control parameter. Substituting \(v\) into Eq. (4.22)
\begin{equation}
\dot{V}_1 = -\lambda_v \rho^2 \cos^2 \alpha \leq 0
\end{equation}

Thus, the designed linear velocity \(v\) ensures that \(\dot{V}_1\) remains non-positive, which leads to the convergence of the position error \(\rho\). However, it is important to note that \(v\) is directly proportional to several variables, including the distance \(\rho\), the target velocity \(v_t\), the cosine of the heading error \(\cos \beta\), and the inverse cosine of the orientation error \(\cos^{-1} \alpha\). These variables tend to have large magnitudes during the initial phase of the system's operation when the robot is far from the target or significantly misaligned. As a result, the computed value of \(v\) may become excessively large at the beginning, which could potentially lead to high control effort or actuator saturation. Therefore, additional considerations or modifications, such as gain tuning or saturation constraints, may be required to ensure the practical feasibility and smooth performance of the controller in real-world implementations.

This issue becomes even more pronounced when designing the angular velocity \(\omega\) to ensure that \(\dot{V}_2\) is non-positive, as both control inputs jointly influence the overall system dynamics. In particular, the dependence of \(\omega\) on variables that also affect \(v\) can exacerbate the problem of excessive control effort in the initial stages, potentially leading to undesirable transient behavior or instability. To mitigate this effect and to prevent the linear velocity from reaching impractically high values, the expression for \(v\) is accordingly modified as follows:
\begin{equation}
v = (v_t \cos \beta + \lambda_v \rho) \cos \alpha
\end{equation}

Substituting into Eq. (4.22) and Eq. (3.14) 
\begin{equation}
\dot{V}_1 = -\lambda_v \rho^2 \cos^2 \alpha + v_t \rho \sin^2 \alpha \cos \beta
\end{equation}

\begin{equation}
\dot{\rho} = -\lambda_v \rho \cos^2 \alpha + v_t \sin^2 \alpha \cos \beta
\end{equation}

From Eq. (4.26), it is clear that \(\dot{V}_1\) will not be non-positive until 
 \(\alpha \to 0 \), the new \(v\) in Eq. (4.25) is designed to reduce the contribution of the target velocity \(v_t\), by multiplying with \(cos \alpha\) and \(cos \beta\), which are bounded by magnitude 1. The contribution of \(\rho\) is attenuated by a factor of \(cos \alpha\). 

For determining the controller input \(\omega\) accordingly, differentiating \(V_2\) with respect to time \(t\) and considering Eq. (3.15), Eq. (3.16) and Eq. (4.25)
\begin{equation}
\begin{aligned}
\dot{V}_2 &= \left( \frac{\sin \alpha}{k_1 \rho} + \frac{\sin \beta}{k_2 \rho} \right)
\bigg[ \left( \frac{\sin 2\alpha}{2} \cos \beta - \sin \beta \right) v_t \\
&\quad {} - \omega \frac{\sin \alpha}{k_1}
- \dot{\phi}_t \frac{\sin \beta}{k_2}
+ \frac{\sin 2\alpha}{2} \lambda_v \left( \frac{\sin \alpha}{k_1} + \frac{\sin \beta}{k_2} \right)
\bigg]
\end{aligned}
\end{equation}

Letting
\begin{equation}
\begin{aligned}
\omega &= \lambda_\alpha \sin \alpha + \left( \frac{\sin \alpha}{k_1 \rho} + \frac{\sin \beta}{k_2 \rho} \right) \bigg[ \left( \frac{k_1 \sin 2\alpha}{2 \sin \alpha} \cos \beta - \frac{k_1 \sin \beta}{\sin \alpha} \right) v_t \\
&\quad {} - \dot{\phi}_t \frac{k_1 \sin \beta}{k_2 \sin \alpha}
+ \frac{k_1 \sin 2\alpha}{2 \sin \alpha} \lambda_v \left( \frac{\sin \alpha}{k_1} + \frac{\sin \beta}{k_2} \right)
\bigg]
\end{aligned}
\end{equation}

Substituting Eq. (4.29) into Eq. (4.28) yields,
\begin{equation}
\dot{V}_2 = -\frac{\lambda_\alpha \sin^2 \alpha}{k_1}
\end{equation}

This shows that \(V_2\) is a non-increasing function of time, since its time derivative \(\dot{V}_2\) is designed to be non-positive. Given \(V_2 > 0\) as defined, \(V_2\) converges to a non-negative limit asymptotically. A similar conclusion holds for \(V_1\), which was previously shown to be non-increasing under the designed control input \(v\). Consequently, both Lyapunov functions \(V_1\) and \(V_2\) remain bounded and approach steady-state values over time. Since \(V_1\) and \(V_2\) are constructed based on the tracking errors in the state variables \(\rho\), \(\alpha\), and \(\beta\), it can be inferred that these variables are also bounded for all \(t \geq 0\). This boundedness ensures that the robot's position and orientation remain within a controlled and predictable range during operation, providing a necessary condition for the overall stability and safety of the tracking control system. Figure 4.1 illustrates the implementation logic for the Lyapunov-based controller.

\begin{figure}[h]
    \begin{center}
        \includegraphics[width=0.8\textwidth]{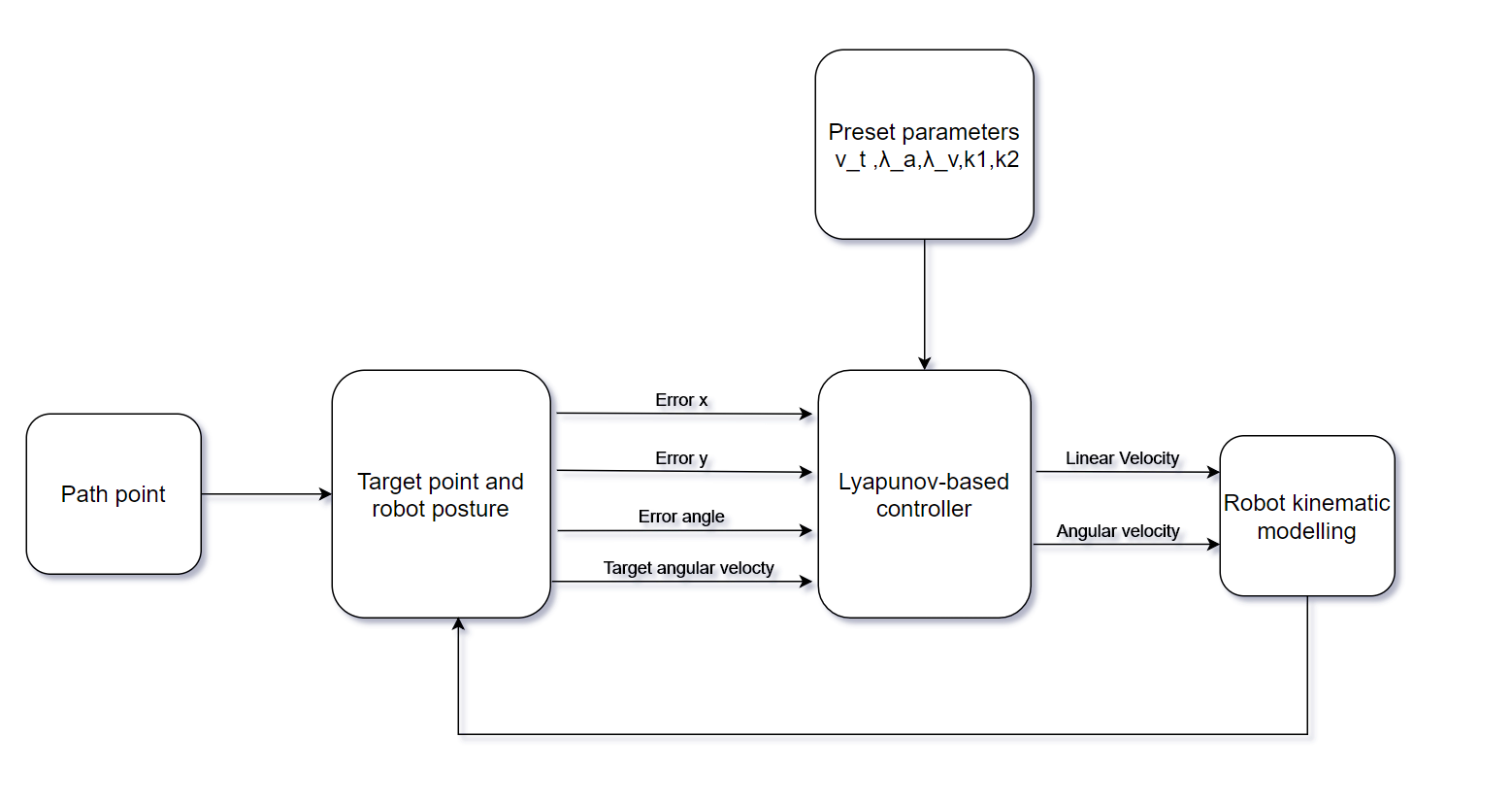}
        \caption{Lyapunuov-based controller diagram}
        \label{Figure_7}
    \end{center}
\end{figure}

Thus, under the action of the designed controller, asymptotic convergence of the tracking errors can be achieved. Specifically, the position error \(\rho\) tends to zero, and the orientation errors \(\alpha\) and \(\beta\) also converge to zero as time progresses, i.e., \(\rho \to 0\), \(\alpha \to 0\), and \(\beta \to 0\) as \(t \to \infty\). This indicates that the robot is able to accurately track the moving target point in both position and orientation. The convergence is guaranteed even when the target point itself is in motion, traveling with a velocity \(v_t\) along a time-varying direction \(\phi_t\). The control strategy effectively compensates for the dynamic nature of the target’s trajectory, ensuring that the robot aligns its motion with the target in real time. As a result, the proposed Lyapunov-based controller not only stabilizes the tracking errors but also enables the robot to follow the moving target smoothly and reliably.

\section{Conventional Lyapunov-based controllers used for comparison}
The performance of the designed controller is evaluated through a comparative analysis with conventional Lyapunov-based controllers, specifically the one proposed in \cite{huang2009control}. In that work, a Lyapunov-based control strategy was developed for an autonomous mobile robot (AMR) to track a moving target point. Similar to the present approach, both the linear and angular velocities, denoted as \(v\) and \(\omega\), are computed using the Lyapunov second method to ensure stability of the tracking process. 

The comparative Lyapunov function of controller is
\begin{align}
V &= V_1 + V_2 \\
V_1 &= \frac{1}{2} \rho^2, \quad V_2 = \frac{1}{2} (\alpha^2 + \beta^2)
\end{align}
In the comparative controller, the Lyapunov function is similarly decomposed into two components, \(V_1\) and \(V_2\), which are individually designed to address the linear and angular tracking errors, respectively. Specifically, \(V_1\) captures the translational error between the robot and the moving target point, while \(V_2\) represents the deviation in orientation.

By choosing appropriate values for \(v\) and \(\omega\), the time derivatives \(\dot{V}_1\) and \(\dot{V}_2\) can be made non-positive, ensuring stability.  \(v\) and \(\omega\) are found to be 
\begin{equation}
v = (v_t \cos \beta + \lambda_v \rho) \cos \alpha
\end{equation}
\begin{equation}
\omega = \lambda_\alpha \alpha + \frac{\alpha + \beta}{\rho} \left( \frac{\sin 2\alpha}{2\alpha} \cos \beta - \frac{\sin \beta}{\alpha} \right) v_t - \frac{\beta}{\alpha} \dot{\phi}_t + \frac{\sin 2\alpha}{2\alpha} \lambda_v (\alpha + \beta)
\end{equation}

\section{Implementation process of autonomous navigation}

The following explains the overall process of robot autonomous navigation. The flow chart of autonomous navigation is shown in Figure 4.2. The following is a brief description of the steps shown in the flow chart.

\begin{enumerate}
    \item Subscribe to the image obtained by the camera.
    \item Detect the lane lines in the image through YOLOP.
    \item Perform polynomial interpolation on the obtained left and right lane lines to generate smooth path points with equal spacing.
    \item Calculate the center line based on the left and right lane lines to obtain the target point.
    \item Based on the target point, calculate the required linear velocity and angular velocity through the controller.
    \item Send the velocity command to the robot kinematic model to achieve the motor rotation.
\end{enumerate}

\begin{figure}[h]
    \begin{center}
        \includegraphics[width=0.4\textwidth]{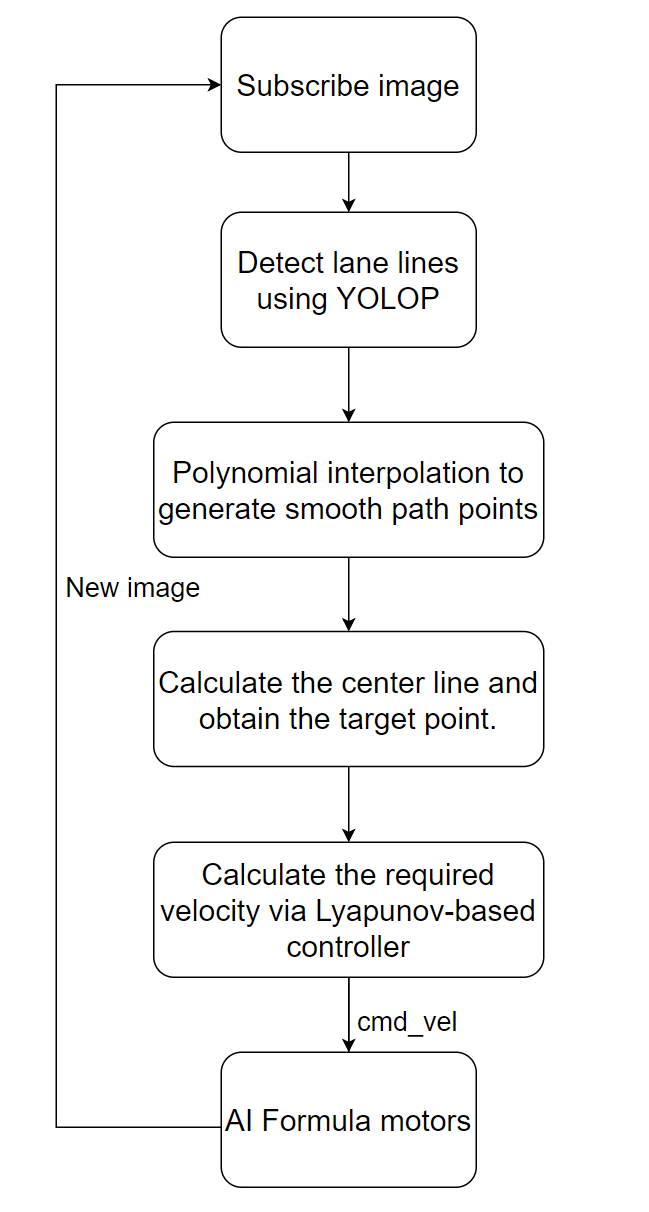}
        \caption{Autonomous navigation flow chart}
        \label{Figure_7}
    \end{center}
\end{figure}

\chapter{Experiment Results and Discussion}

\section{Experimental scenario}
To further validate the practical applicability of the proposed autonomous navigation strategy in a race track environment, real-world experiments were conducted. The evaluation was carried out using the AI Formula unmanned vehicle, as introduced in Chapter 1, which was controlled by the Lyapunov-based control system developed in this study. The main hardware and software configurations used in the experimental setup are detailed as Table. 5.1.

\begin{table}[ht]
\centering
\caption{Hardware and software configuration}
\begin{tabular}{|c|l|c|}
\hline
\textbf{Hardware and software} & \textbf{Model/Specification} \\
\hline
PC & Jetson AGX Orin \\
\hline
OS & Ubuntu 20.04 \\
\hline
ROS version & ROS Foxy \\
\hline
Camera & ZEDX \\
\hline
GNSS+IMU & VN200 \\
\hline
Wheel encoder & FBLG2360T \\
\hline
\end{tabular}
\end{table}

The experimental tests were conducted on the race track described in Chapter 1. A complete lap of autonomous navigation was performed at varying speeds, starting line, finishing line and different scenarios shown in Figures 5.2 to 5.6 at the locations indicated in Figure 5.1. The experimental data collected under these conditions were subsequently compared and analyzed. Based on prior experience with parameter tuning during preliminary experiments, the control gains were set to \(\lambda_v = 0.075\) and \(\lambda_a = 0.15\). For the proposed controller, the parameters were configured as \(k_1 = 0.8\) and \(k_2 = 50\).

\begin{figure}[h]
    \begin{center}
        \includegraphics[width=0.6\textwidth]{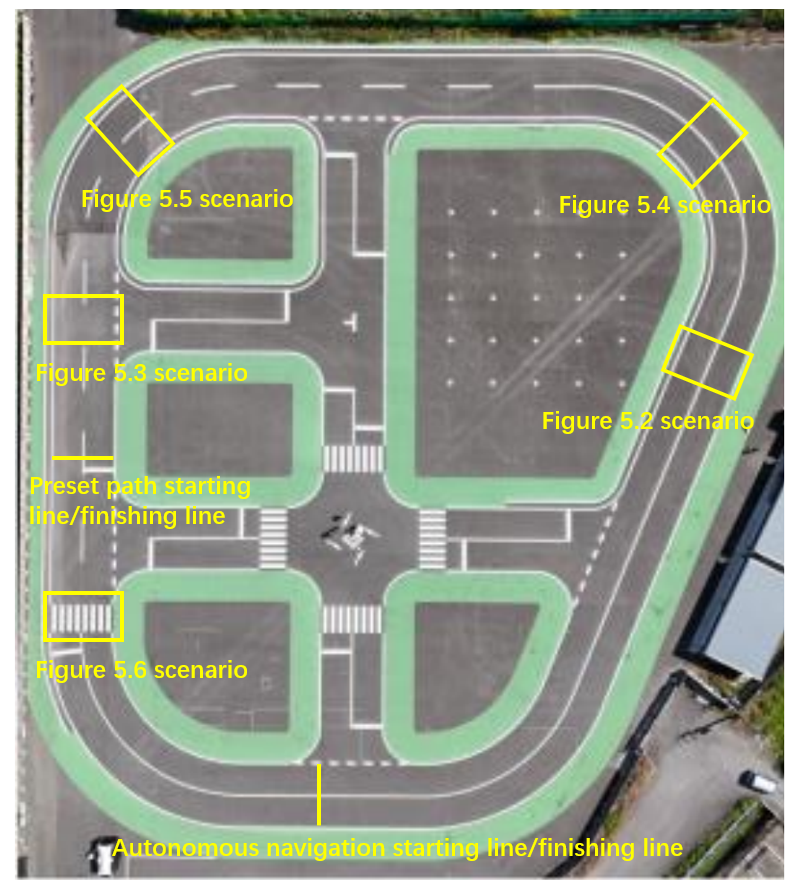}
        \caption{Top view and starting line, finishing line and different scenarios of the race track}
        \label{Figure_7}
    \end{center}
\end{figure}

In addition, to ensure operational safety and maintain the reliability of visual perception during autonomous navigation, upper and lower bounds were imposed on both the linear and angular velocities of the robot. These constraints are particularly important to mitigate the risk of recognition failures caused by direct sunlight interfering with the stereo camera's field of view during operation. By limiting the robot's motion within predefined velocity ranges, the system can avoid excessive speed that may lead to unstable behavior.

\begin{align}
& 0.6 \leq v \leq 1.75, \text{when} \, v_t = 1.5 m/s \\
& 0.6 \leq v \leq 2.25, \text{when} \,v_t = 2.0 m/s \\
& -0.4 \leq \omega \leq 0.4 
\end{align}

\section{Perception results}
This section primarily presents the recognition results obtained using the YOLOP model, along with the corresponding outcomes of the polynomial curve fitting process applied to the detected lane lines. The purpose of this analysis is to evaluate the accuracy and reliability of the visual perception module in extracting lane features and generating smooth, continuous representations of the driving path, which are essential for subsequent path planning and control. It should be noted that the centers in the following grid maps are the geometric centers of the robot's kinematic model, and the direction in which the robot moves is recorded as the x direction. Each subplot in Section 5.2, the green dotted line corresponds to the obtained lane line, while the blue dotted line represents the center line calculated by obtained lane lines.

Figure 5.2 illustrates the scenario in which the lane line captured in the image are straight, solid lines. As shown, the recognition performance is highly satisfactory. The lane line detected by the YOLOP and the corresponding centerline generated through polynomial fitting closely align with expectations. The resulting fitted curves are smooth and accurate, thereby providing reliable and high-quality input for the controller in the autonomous navigation system.
\begin{figure}[h]
    \begin{center}
        \includegraphics[width=0.75\textwidth]{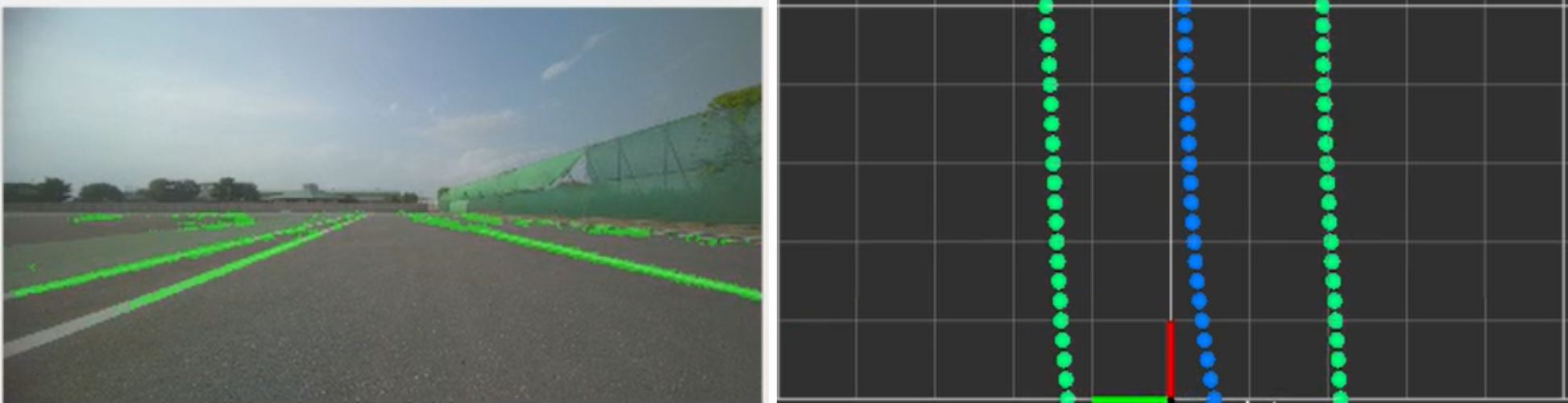}
        \caption{Perception and fitting results of straight and solid lines}
        \label{Figure_7}
    \end{center}
\end{figure}

Figure 5.3 depicts the scenario in which the lane lines in the image are straight and consist of dotted and solid lines. It can be observed that the recognition results differ to some extent from the lane lines and centerline generated through polynomial fitting. This discrepancy is primarily attributed to the intermittent gaps in the dotted lines, which introduce deviations in the fitting results from the actual lane geometry. Nevertheless, in the regions where the white line segments are present, both the detection and the fitted curves are consistent with expectations, demonstrating reliable performance in those segments.
\begin{figure}[h]
    \begin{center}
        \includegraphics[width=0.75\textwidth]{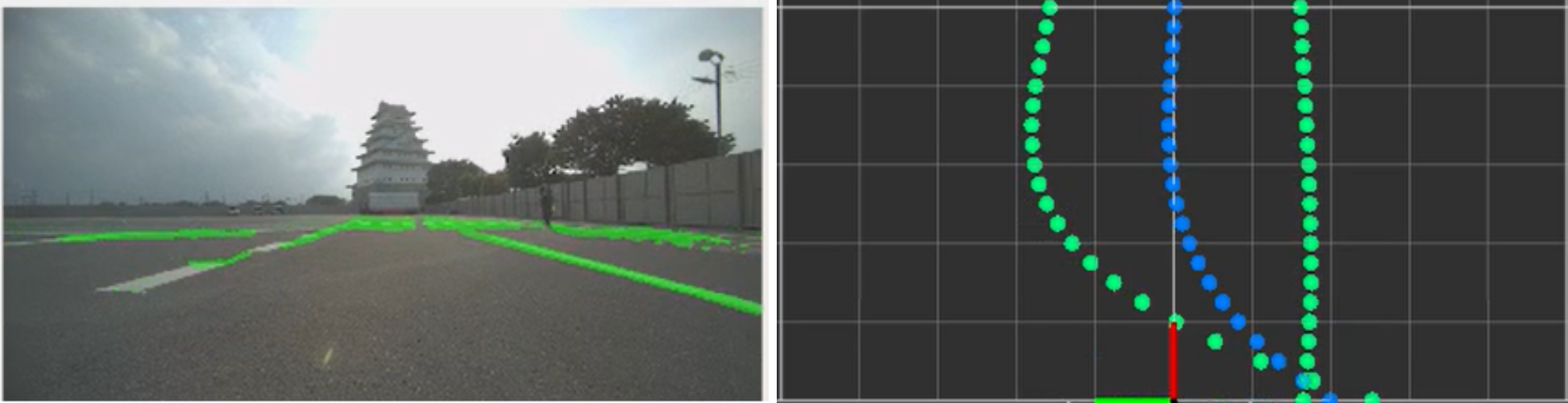}
        \caption{Perception and fitting results of straight and dotted lines}
        \label{Figure_7}
    \end{center}
\end{figure}

Figure 5.4 presents the scenario in which the lane lines in the image are both curve and solid. As illustrated, the recognition results are highly satisfactory. The detected lane line and the corresponding centerline obtained through polynomial fitting closely align with the expected geometry. The fitted curves are smooth and accurate, thereby offering high-quality input to the controller for effective path tracking.
\begin{figure}[h]
    \begin{center}
        \includegraphics[width=0.75\textwidth]{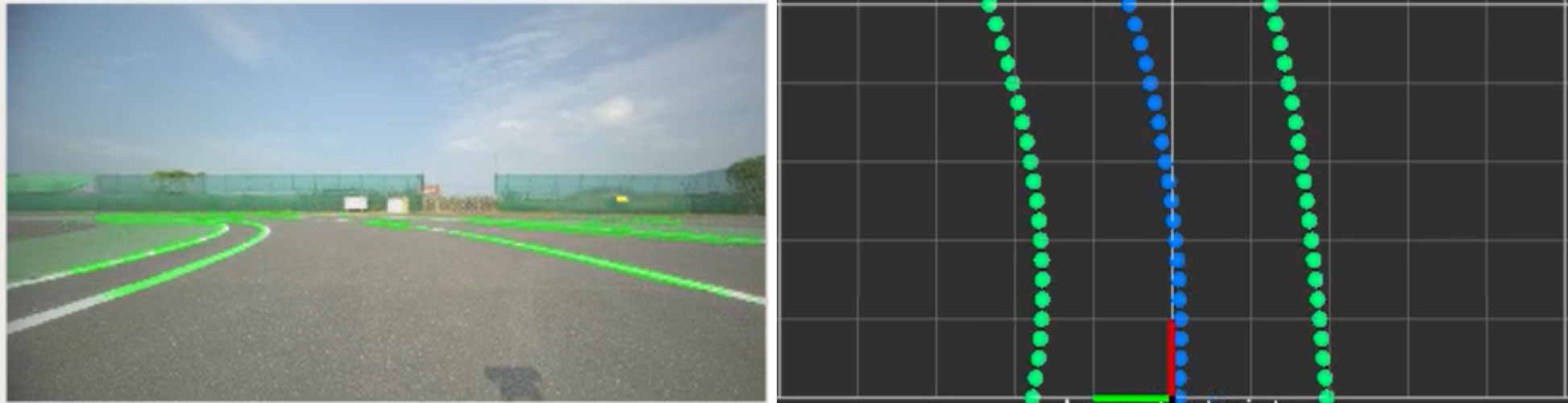}
        \caption{Perception and fitting results of curve and solid lines}
        \label{Figure_7}
    \end{center}
\end{figure}

Figure 5.5 illustrates the scenario in which the lane lines in the image are curve and consist of dotted and straight line. It is evident that the recognition performance on the side with the dotted line, as well as the polynomial fitting results for the lane line and the resulting centerline, are suboptimal. This reduced accuracy is primarily due to the presence of intermittent gaps in the dotted line, which cause the fitting process to deviate from the true lane geometry. Consequently, both the recognition and fitting outcomes do not fully meet expectations in this case.
\begin{figure}[h]
    \begin{center}
        \includegraphics[width=0.75\textwidth]{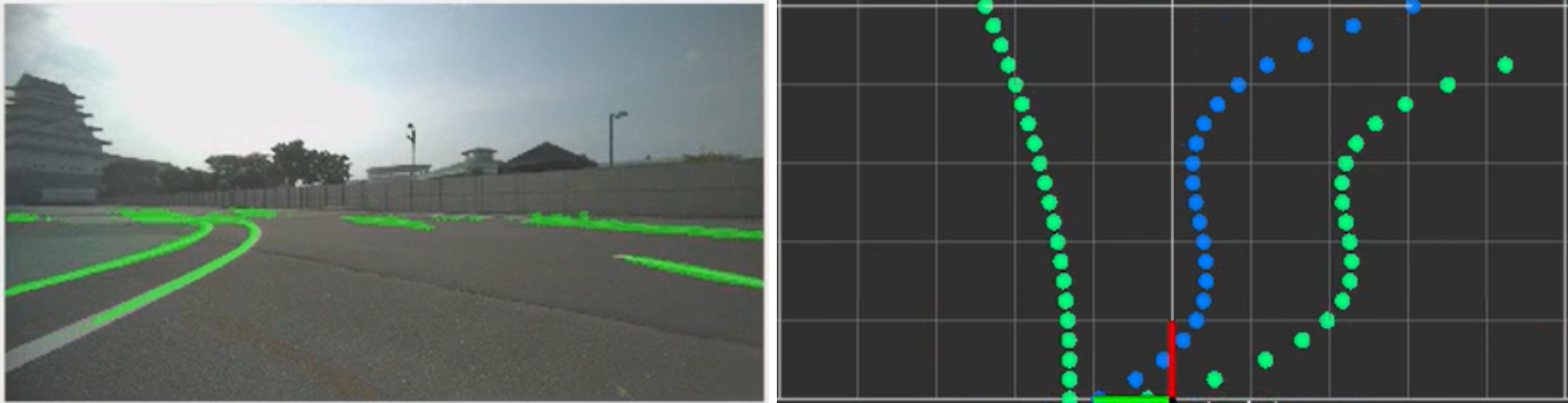}
        \caption{Perception and fitting results of curve and dotted lines}
        \label{Figure_7}
    \end{center}
\end{figure}

Figure 5.6 depicts the scenario in which the image contains a zebra crossing. As observed, the recognition performance in this case is significantly degraded. The presence of dense and irregular patterns interferes with lane line detection, resulting in a distorted polynomial fitting outcome. Consequently, a reliable and continuous centerline cannot be generated, leading to a failure in providing stable input to the controller at that moment.
\begin{figure}[h]
    \begin{center}
        \includegraphics[width=0.75\textwidth]{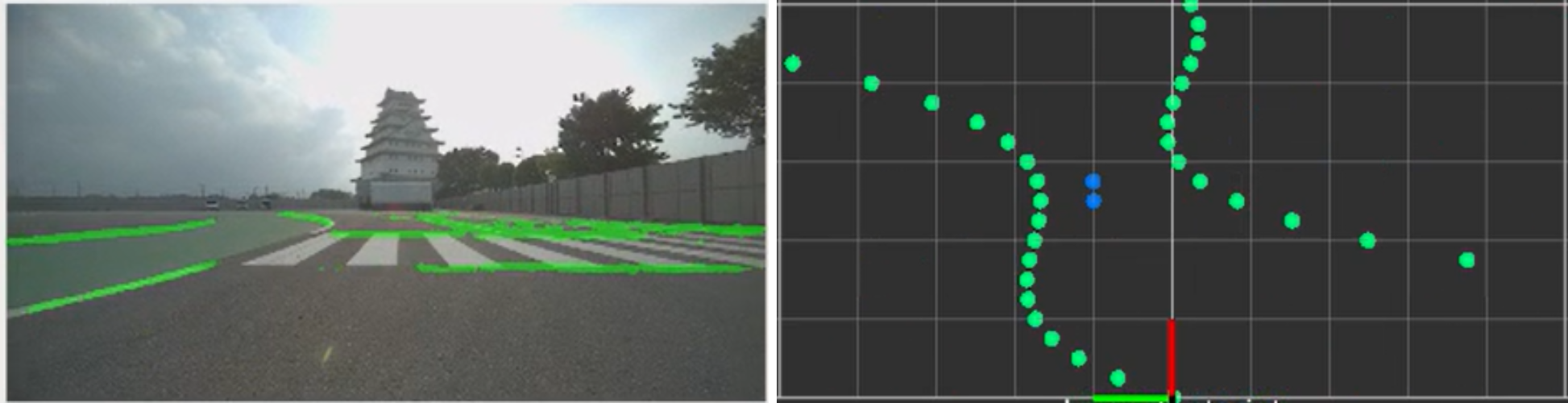}
        \caption{Perception and fitting results of zebra crossing}
        \label{Figure_7}
    \end{center}
\end{figure}

Although the perception and fitting results in the five aforementioned images vary in quality, it is important to note that the controller’s input primarily relies on three look ahead points. From the visualizations, it can be observed that at a distance of approximately 2–3 meters from the vehicle's coordinate origin (with each grid unit in the figure corresponding to 0.5 × 0.5 meters in the real world), the perception outcomes—regardless of whether the lane is straight or curved, solid or dotted—tend to remain stable and accurately reflect the expected centerline. Based on this observation, a point located 2 meters ahead of the vehicle center along the x-axis is selected as the primary target point. Additionally, two more points spaced at 0.5-meter intervals in the positive x-axis from the target point are chosen to form a set of three look ahead points. This configuration ensures that the controller receives consistent and reliable input during the robot's autonomous navigation process.

\section{Preset path based tracking results}

This section primarily presents the results of the robot's path tracking performance under the control of the Lyapunov-based controller based on preset path. The objective is to evaluate the effectiveness and stability of this control strategy in guiding the robot along a given path, independent of real-time visual perception. By analyzing tracking accuracy and motion behavior, the results aim to validate the controller’s capability to achieve smooth and reliable tracking effect under varying target speed conditions. In each subplot in Section 5.3, the blue curve corresponds to the trajectory-following control method, while the red curve represents the results obtained using the comparative method.

Figure 5.7 illustrates the experimental comparison of path tracking performance between the two control methods at $v_t=1.5m/s$. It is evident that the trajectory of proposed Lyapunov-based controller, exhibits a smooth and stable path throughout the entire course. In contrast, the comparative method fails to effectively follow the preset path and, in some segments, deviates significantly from the reference path. The robot controlled by the proposed strategy maintains accurate tracking with producing a smoother driving trajectory.. This superior performance highlights the robustness and precision of the proposed method in executing reliable and continuous path tracking, thereby demonstrating its effectiveness in autonomous navigation tasks based on preset path.
\begin{figure}[h]
    \begin{center}
        \includegraphics[width=0.85\textwidth]{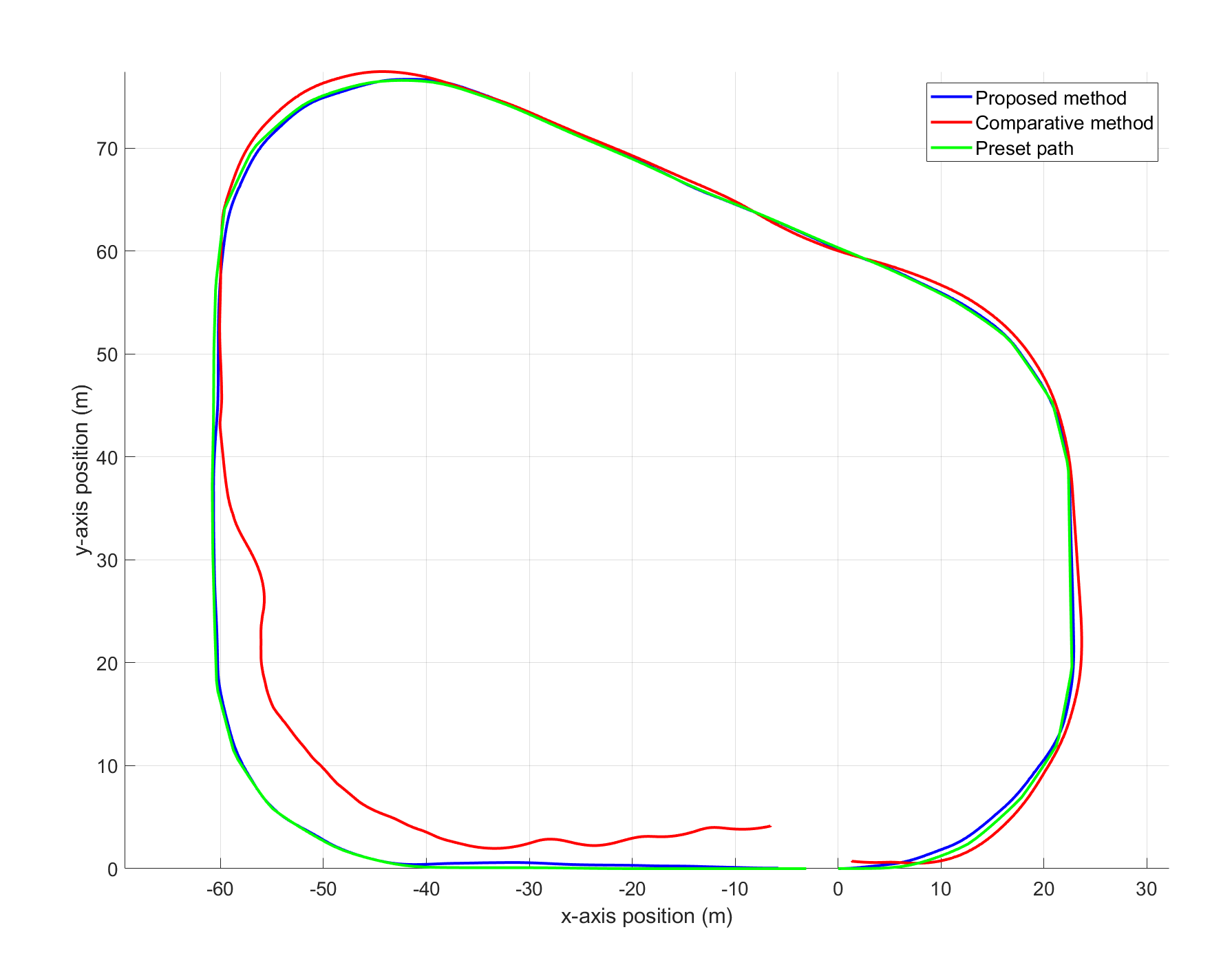}
        \caption{Trajectory results of the two controllers when the target speed \(v_t = 1.5 m/s\)}
        \label{Figure_7}
    \end{center}
\end{figure}

Figure 5.10, from top to bottom, displays the time-series data of the robot's $v$, $\omega$, $x$ along the x-axis, $y$ along the y-axis, and $\phi$ when $v_t=1.5 m/s$. It is evident that the proposed method results in significantly smoother and more stable control performance compared to the comparative method. The linear speed remains close to the target with minimal fluctuations, whereas the comparative method exhibits multiple sharp drops and large deviations, even triggering velocity and acceleration limit constraints during execution. Similarly, the angular velocity under the proposed method maintains lower variance, indicating more consistent heading adjustments. In the positional plots along both the \(x\)- and \(y\)-axes, the proposed method follows a continuous and predictable trajectory, while the comparative method shows larger deviations, particularly around the middle and later segments of the trajectory. Notably, the orientation \(\phi\) remains stable throughout the entire path with the proposed controller, whereas the comparative method introduces sudden jumps and discontinuities, indicating control instability. These results collectively validate the superior tracking accuracy, dynamic smoothness, and control robustness of the proposed Lyapunov-based method under moderate-speed navigation.
\begin{figure}[h]
    \begin{center}
        \includegraphics[width=0.9\textwidth]{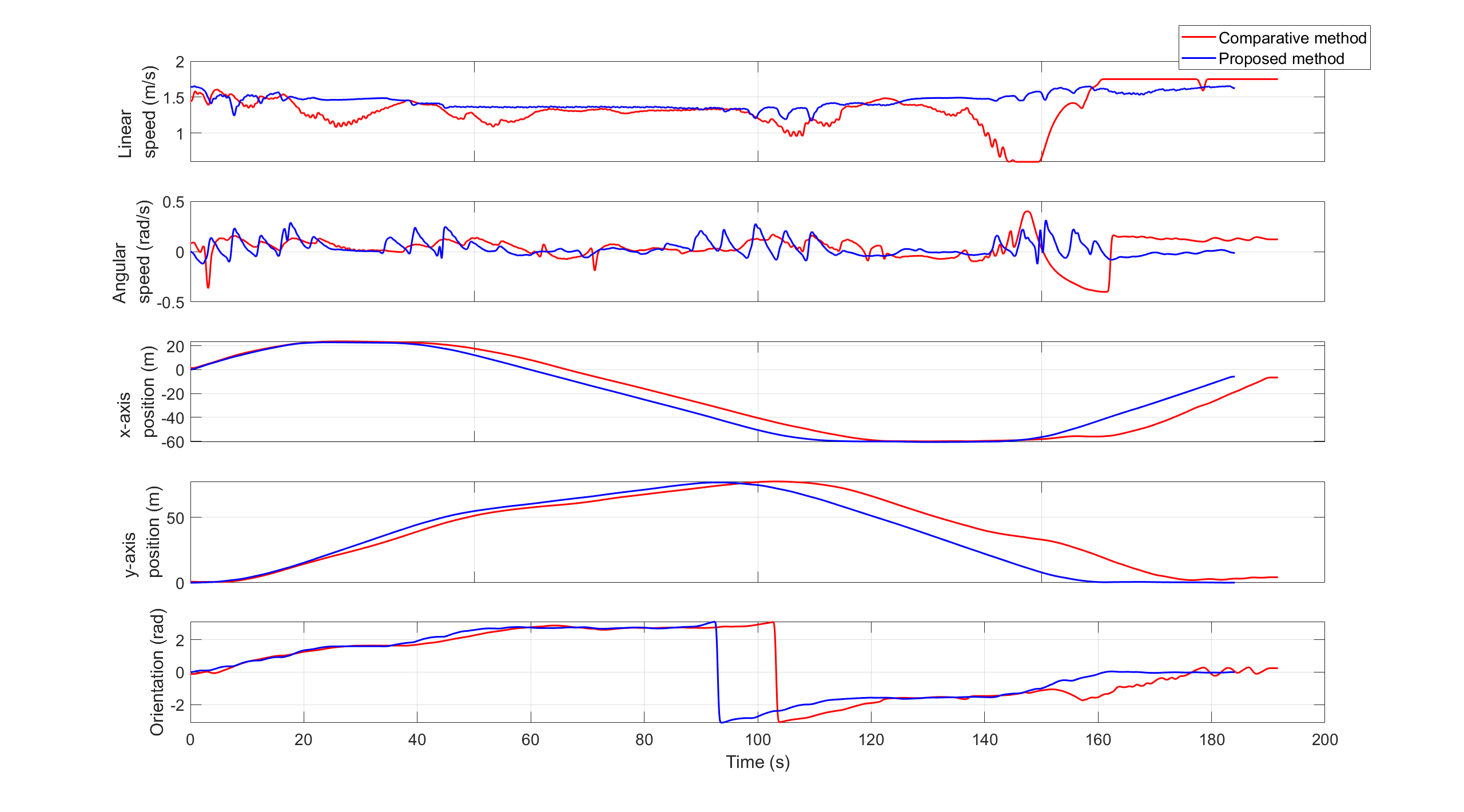}
        \caption{Results with two controllers when the target speed \(v_t = 1.5 m/s\), from top to bottom: linear speed, angular speed, x-axis position, y-axis position, and the orientation}
        \label{Figure_7}
    \end{center}
\end{figure}

Table 5.2 provides a detailed quantitative comparison of the proposed method and the comparative method across several performance metrics under the condition of a target speed \(v_t = 1.5\,\text{m/s}\). The experimental results demonstrate the clear superiority of the proposed Lyapunov-based control strategy in terms of tracking accuracy, control stability, and overall efficiency. In terms of task completion, the proposed method completes the course in 183.60\,s, which is 8.1\,s faster than the comparative method, while simultaneously achieving a higher average linear speed. This indicates that the proposed controller is capable of maintaining higher velocity without compromising path stability. From a path tracking perspective, the proposed method yields a significantly lower mean absolute error(MAE) of lateral compared to the comparative method, as well as a substantial reduction in the mean MAE of orientation. These results confirm improved lateral precision and heading consistency during navigation. In terms of target speed, the proposed controller demonstrates a much lower root mean square error (RMSE) of linear speed compared to the comparative method. Furthermore, the linear speed deviation is reduced by more than two-thirds, reflecting more stable and accurate speed control. Finally, the accumulated orientation is considerably lower for the proposed method compared to the comparative method, indicating smoother directional changes and better dynamic stability. 
\begin{table}[h]
\centering
\caption{Performance metrics comparison between proposed and comparative methods when the target speed \(v_t = 1.5 m/s\)}
\begin{tabular}{|l|c|c|}
\hline
 & Proposed method & Comparative method \\
\hline
Completion time (s) & 183.60 & 191.70 \\
\hline
Average Linear speed (m/s) & 1.45 & 1.35 \\
\hline
Average angular speed (rad/s) & 0.032 & 0.035 \\
\hline
MAE of lateral (m) & 0.20 & 1.13 \\
\hline
MAE of orientation (rad) & 0.069 & 0.31 \\
\hline
RMSE of linear speed (m/s) & 0.014 & 0.090 \\
\hline
Linear speed deviation (\%) & 3.33 & 10.00 \\
\hline
Accumulated orientation (rad) & 7.43 & 19.71 \\
\hline
\end{tabular}
\end{table}

Overall, these results quantitatively validate the effectiveness and robustness of the proposed trajectory-following control approach. It not only enables more accurate and stable path tracking but also enhances motion efficiency and smoothness, outperforming the comparative method across all evaluated metrics.

Figure 5.9 presents the experimental comparison of path tracking performance between the proposed Lyapunov-based controller and the comparative method at a target speed of \(v_t = 2.0\,\text{m/s}\). As shown in the figure, the trajectory generated by the proposed method closely follows the preset path  throughout the entire course, maintaining high consistency and smoothness. In contrast, the comparative method exhibits a substantial deviation from the preset path in the latter segment of the trajectory, where it fails to complete the loop and diverges significantly from the intended route.
\begin{figure}[h]
    \begin{center}
        \includegraphics[width=0.85\textwidth]{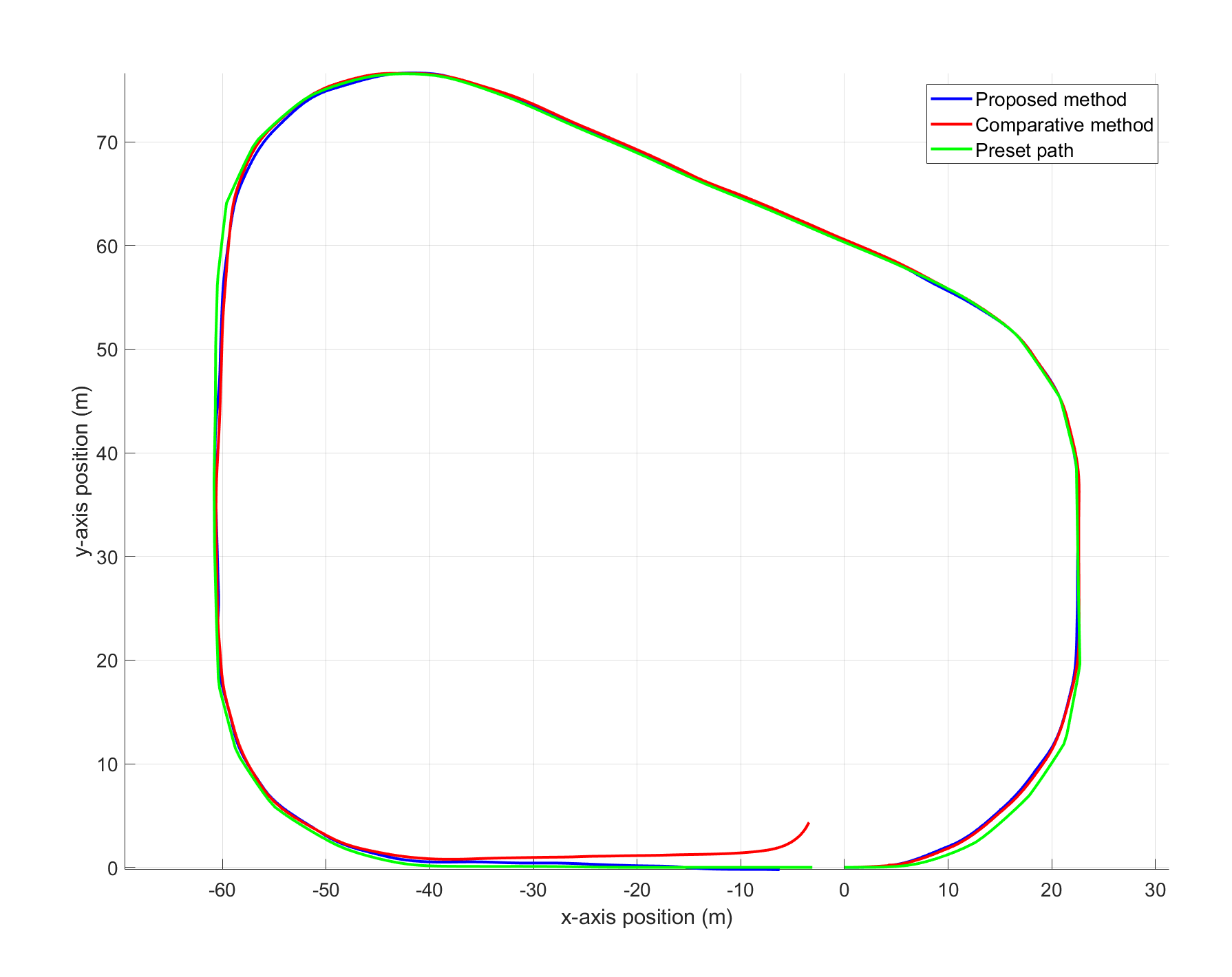}
        \caption{Trajectory results of the two controllers when the target speed \(v_t = 2.0 m/s\)}
        \label{Figure_7}
    \end{center}
\end{figure}

Figure 5.10, from top to bottom, displays the time-series data of the robot's $v$, $\omega$, $x$ along the x-axis, $y$ along the y-axis, and $\phi$ when $v_t=2.0 m/s$. The proposed method exhibits more stable and consistent control performance compared to the comparative method . The linear speed remains close to the target speed with minimal fluctuation throughout the trajectory, whereas the comparative method shows significant speed degradation in the later stage, indicating loss of velocity tracking. Similarly, the angular velocity under the proposed controller remains well-regulated, with reduced oscillation amplitude, while the comparative method exhibits pronounced instability, particularly in the final segment, where excessive fluctuations and sharp peaks occur. In terms of position tracking along both the \(x\)- and \(y\)-axes, the trajectories of the proposed method align closely with the reference path, while the comparative method demonstrates increasing deviation as time progresses. The orientation profile \(\phi\) further confirms the superior performance of the proposed method. The heading angle changes smoothly without abrupt transitions, in contrast to the comparative method, which suffers from a sudden and unstable orientation reversal near the 70–80\,s mark. These results collectively highlight the robustness, accuracy, and dynamic stability of the proposed Lyapunov-based controller in maintaining reliable path tracking at higher speeds.
\begin{figure}[h]
    \begin{center}
        \includegraphics[width=0.9\textwidth]{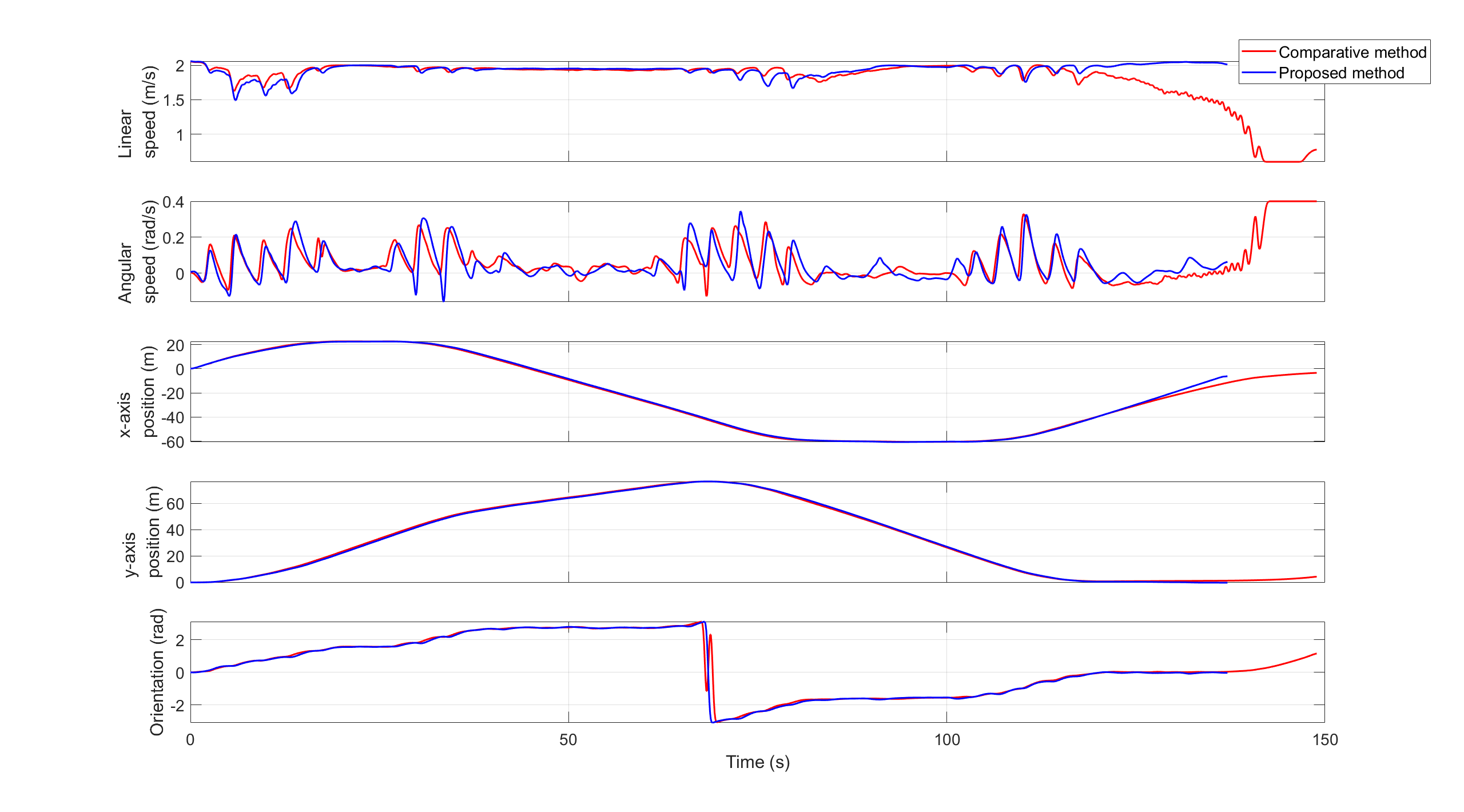}
        \caption{Results with two controllers when the target speed \(v_t = 2.0 m/s\), from top to bottom: linear speed, angular speed, x-axis position, y-axis position, and the orientation}
        \label{Figure_7}
    \end{center}
\end{figure}

Table 5.3 presents a quantitative comparison of the proposed method and the comparative method under the target speed \(v_t = 2.0\,\text{m/s}\), across a range of performance evaluation metrics. The results demonstrate that the proposed Lyapunov-based control strategy maintains its superior performance, even under higher-speed conditions. In terms of efficiency, the proposed method achieves a shorter completion time, while also attaining a higher average linear speed. These improvements indicate that the proposed controller allows the robot to complete the task more quickly while preserving trajectory integrity. From the perspective of path tracking, the proposed method exhibits better performance, with a lower MAE of lateral and a reduced MAE of orientation. Although the tracking errors are slightly higher than those observed at the lower speed of \(1.5\,\text{m/s}\), the overall accuracy remains within acceptable bounds and superior to the baseline. Regarding speed control, the proposed method achieves a significantly lower RMSE of linear speed compared to the comparative method, and a notably smaller linear speed deviation , reflecting consistent and stable velocity regulation even at elevated speeds. Finally, the accumulated orientation of the proposed method is slightly lower than that of the comparative method, indicating smoother turning behavior and better directional stability. Notably, the accumulated orientation at \(2.0\,\text{m/s}\) remains close to that observed at \(1.5\,\text{m/s}\), suggesting that the proposed controller successfully maintains its stability across different speed regimes.
\begin{table}[ht]
\centering
\caption{Performance metrics comparison between proposed and comparative methods when the target speed \(v_t = 2.0 m/s\)}
\begin{tabular}{|l|c|c|}
\hline
 & Proposed method & Comparative method \\
\hline
Completion time (s) & 136.60 & 148.90 \\
\hline
Average linear speed (m/s) & 1.93 & 1.80 \\
\hline
Average angular speed (rad/s) & 0.042 & 0.062 \\
\hline
MAE of lateral (m) & 0.26 & 0.49 \\
\hline
MAE of orientation (rad) & 0.079 & 0.11 \\
\hline
RMSE of linear speed (m/s) & 0.013 & 0.15 \\
\hline
Linear speed deviation (\%) & 3.50 & 10.00 \\
\hline
Accumulated orientation (rad) & 7.83 & 8.22 \\
\hline
\end{tabular}
\end{table}

Overall, these results affirm the robustness and adaptability of the proposed trajectory-following control framework. It continues to outperform the comparative method across all evaluated metrics, even under increased dynamic complexity associated with higher target speeds.

\section{Vision-based path tracking results}
This section primarily presents the results of the robot's path tracking performance under the control of the Lyapunov-based controller. The objective is to evaluate the effectiveness and stability of the proposed control strategy in guiding the robot along the planned trajectory. By analyzing the tracking accuracy and overall motion behavior, the results serve to validate the controller’s ability to ensure smooth and reliable autonomous navigation under varying target speed conditions. Each subplot in Section 5.4, the blue curve corresponds to the proposed Lyapunov-based control method, while the red curve represents the results obtained using the comparative method.

Figure 5.11 illustrates the experimental comparison of path tracking performance between the two control methods at $v_t=1.5m/s$. It is clearly observable that the trajectory of proposed Lyapunov-based controller, exhibits a smoother and more stable path compared to the comparative method. The robot following the proposed control strategy demonstrates reduced lateral oscillations and minimizes unnecessary rotational movements while approaching and tracking the target points. This improved behavior indicates enhanced control precision and better dynamic stability, thereby validating the effectiveness of the proposed method in achieving more efficient and reliable autonomous navigation.
\begin{figure}[h]
    \begin{center}
        \includegraphics[width=0.75\textwidth]{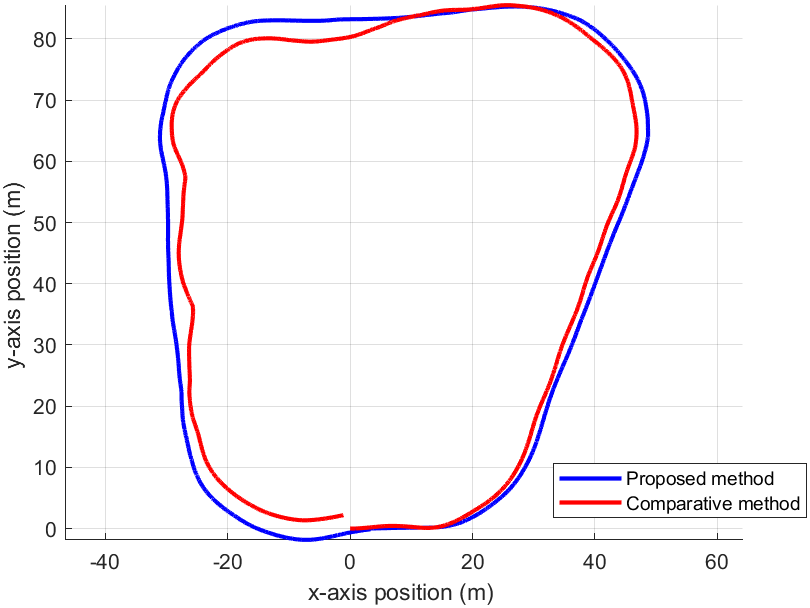}
        \caption{Trajectory results of the two controllers when the target speed \(v_t = 1.5 m/s\)}
        \label{Figure_7}
    \end{center}
\end{figure}

Figure 5.12, from top to bottom, displays the time-series data of the robot's $v$, $\omega$, $x$ along the x-axis, $y$ along the y-axis, and $\phi$ when $v_t=1.5 m/s$. It can be observed that the proposed method achieves more stable control behavior, as reflected by smoother profiles in both linear and angular speeds. Furthermore, the frequency of triggering the lower bound of linear speed as well as the upper and lower bounds of acceleration is noticeably reduced compared to the comparative method. These limit-triggering events primarily occur when the robot passes through regions with frequent interruptions in visual cues, such as areas with concentrated dotted lines and zebra crossings in Figure 5.2, which lead to unstable target point inputs.
\begin{figure}[h]
    \begin{center}
        \includegraphics[width=0.9\textwidth]{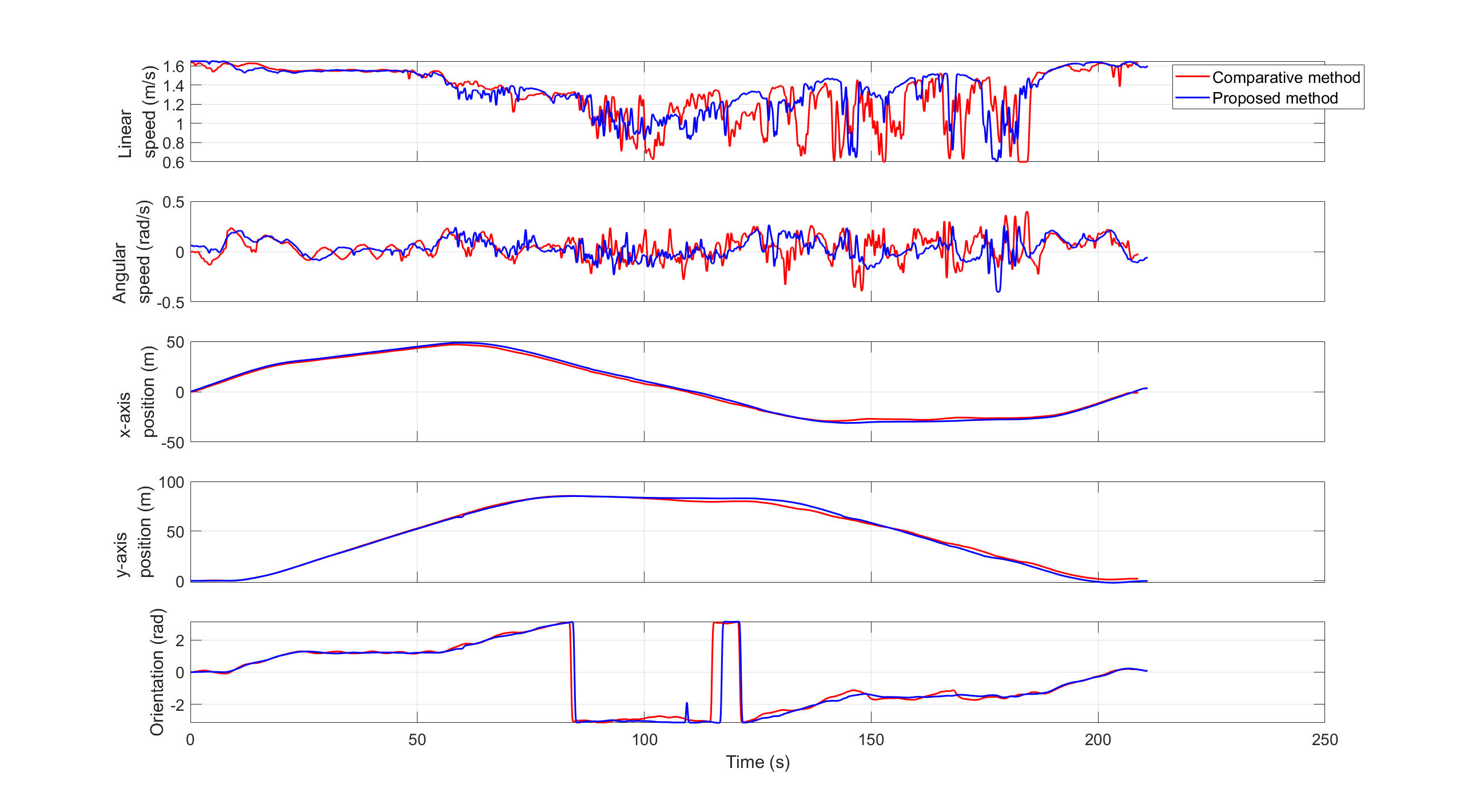}
        \caption{Results with two controllers when the target speed \(v_t = 1.5 m/s\), from top to bottom: linear speed, angular speed, x-axis position, y-axis position, and the orientation}
        \label{Figure_7}
    \end{center}
\end{figure}

Table 5.4 presents a quantitative comparison between the proposed method and the comparative method across a range of evaluation metrics. The results clearly demonstrate that the proposed control strategy achieves superior performance in multiple aspects. Specifically, it achieves a slightly shorter completion time and a higher average linear speed , indicating improved efficiency. Moreover, the proposed method exhibits a significantly lower MAE in lateral positioning and orientation , reflecting enhanced tracking accuracy and directional stability. In terms of speed control, it also achieves lower RMSE of linear speed and reduced linear speed deviation with target speed, which indicates more stable velocity regulation. Finally, the accumulated orientation is lower, indicating that the proposed method enables smoother and more efficient path following. Collectively, these results confirm the effectiveness and robustness of the proposed control approach over the comparative method.
\begin{table}[h]
\centering
\caption{Performance metrics comparison between proposed and comparative methods when the target speed \(v_t = 1.5 m/s\)}
\begin{tabular}{|l|c|c|}
\hline
 & Proposed method & Comparative method \\
\hline
Completion time (s) & 206.54 & 207.00 \\
\hline
Average Linear speed (m/s) & 1.35 & 1.31 \\
\hline
Average angular speed (rad/s) & 0.037 & 0.038 \\
\hline
MAE of lateral (m) & 0.51 & 0.61 \\
\hline
MAE of orientation (rad) & 0.15 & 0.23 \\
\hline
RMSE of linear speed (m/s) & 0.091 & 0.13 \\
\hline
Linear speed deviation (\%) & 10.00 & 12.67 \\
\hline
Accumulated orientation (rad) & 20.36 & 26.02 \\
\hline
\end{tabular}
\end{table}

Figure 5.13 presents an experimental comparison of the trajectory tracking performance of the two control strategies when $v_t=2.0m/s$. It is evident that the trajectory of the proposed Lyapunov-based controller continues to demonstrate a smoother and more stable path compared to the comparative method. The robot operating under the proposed control scheme exhibits reduced lateral oscillations and minimized unnecessary rotational motions while tracking the designated target points. Nevertheless, it is noted that the overall smoothness of the trajectory is slightly diminished relative to the results observed at lower speeds, reflecting the increased difficulty of maintaining optimal tracking performance as speed increases.
\begin{figure}[h]
    \begin{center}
        \includegraphics[width=0.75\textwidth]{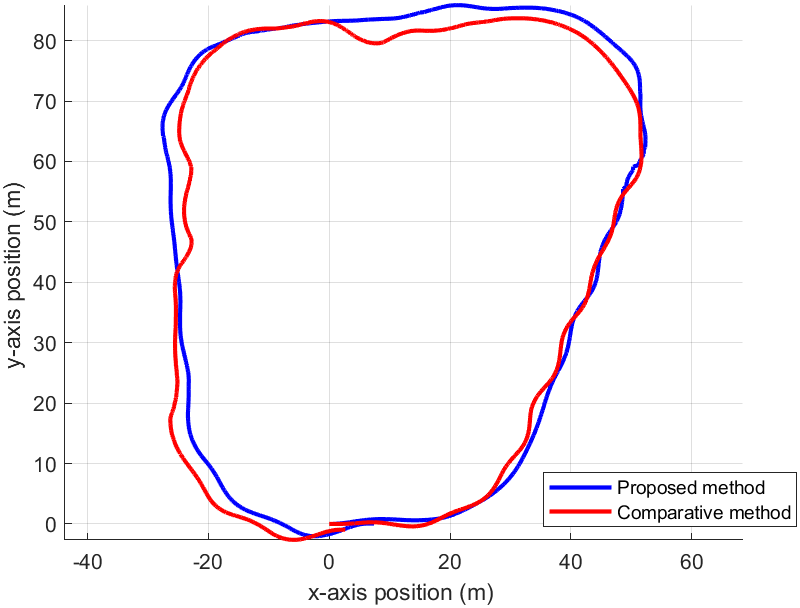}
        \caption{Trajectory results of the two controllers when the target speed \(v_t = 2.0 m/s\)}
        \label{Figure_7}
    \end{center}
\end{figure}

Figure 5.14, from top to bottom, displays the time-series data of the robot's $v$, $\omega$, $x$ along the x-axis, $y$ along the y-axis, and $\phi$ when $v_t=2.0m/s$.  It can be observed that the angular speed of the robot under the baseline control method exhibits a significantly wider range of variation, indicating less stable motion and poorer trajectory tracking performance. In contrast, the proposed Lyapunov-based controller demonstrates a more consistent and constrained angular velocity profile, reflecting its superior path tracking capability and enhanced stability during navigation. This reduction in angular spped fluctuations suggests that the proposed method enables smoother directional adjustments. However, it is also noted that the overall control performance is somewhat diminished compared to the results obtained at lower speeds. This performance degradation is likely attributable to the increased dynamic complexity and reduced response time associated with higher target velocities, which pose greater challenges for maintaining precise control.
\begin{figure}[h]
    \begin{center}
        \includegraphics[width=0.9\textwidth]{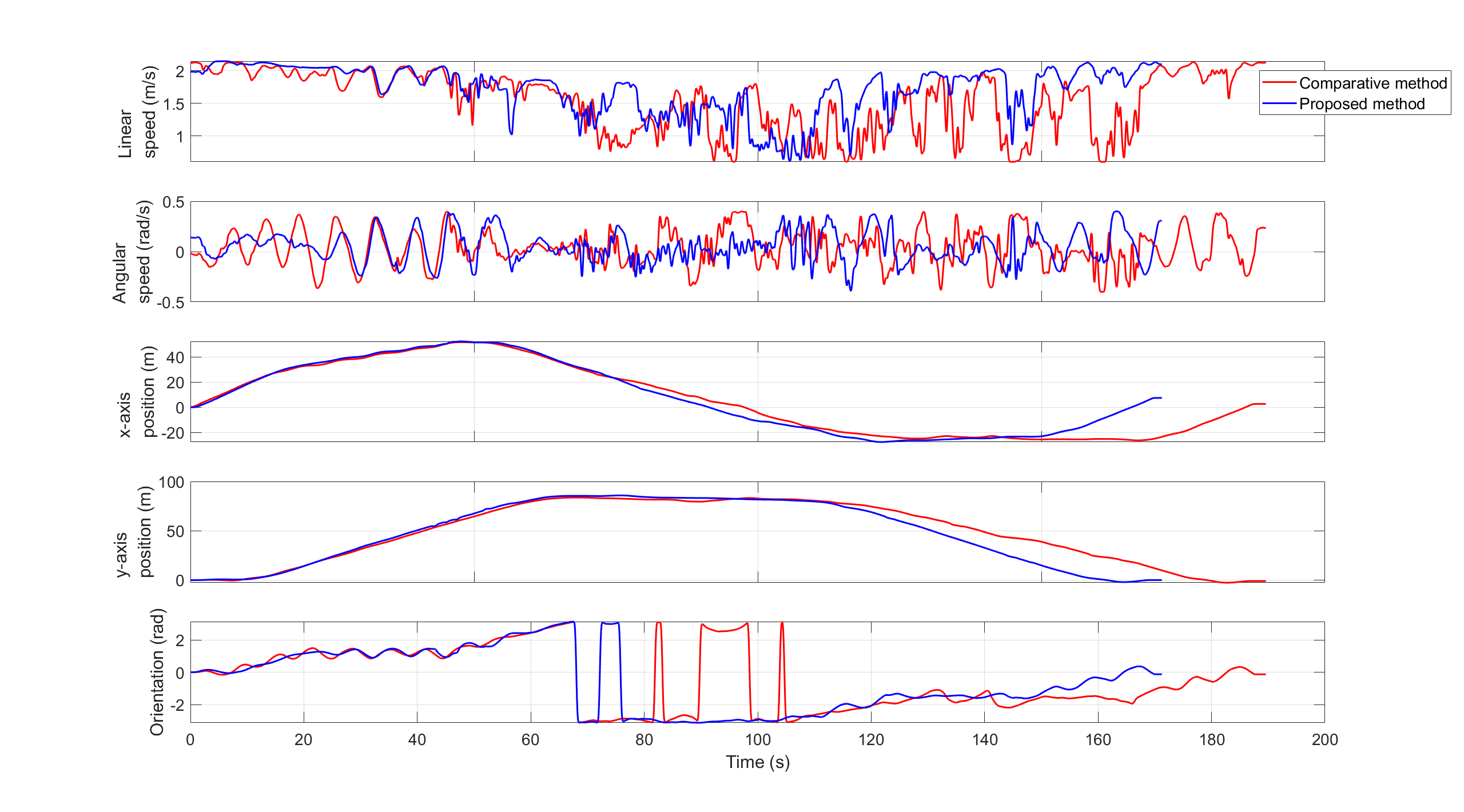}
        \caption{Results with two controllers when the target speed \(v_t = 2.0 m/s\), from top to bottom: linear speed, angular speed, x-axis position, y-axis position, and the orientation}
        \label{Figure_7}
    \end{center}
\end{figure}

Table 5.5 presents a comprehensive quantitative comparison between the proposed method and the comparative method across multiple evaluation metrics when $v_t=2.0m/s$. The proposed method demonstrates overall superior performance. It achieves a marginally shorter completion time and a higher average linear speed, indicating improved operational efficiency. Although the average angular speed values are similar, significant advantages are observed in tracking accuracy. Specifically, the proposed method yields a substantially MAE of lateral and MAE of orientation. Additionally, it exhibits a RMSE in linear speed and a smaller linear speed deviation, which reflect enhanced stability in velocity control. Furthermore, the accumulated orientation is also lower for the proposed method, suggesting smoother heading transitions throughout the trajectory. Collectively, these results underscore the effectiveness, accuracy, and robustness of the proposed Lyapunov-based control strategy under high-speed autonomous navigation conditions. However, it is important to note that, overall, all the aforementioned evaluation metrics exhibit a decline in performance when compared to those obtained under lower-speed conditions. As a result, while the proposed method still outperforms the comparative method at higher speeds, its absolute performance across metrics such as tracking accuracy, velocity stability, and orientation control is diminished relative to its performance in low-speed scenarios.
\begin{table}[ht]
\centering
\caption{Performance metrics comparison between proposed and comparative methods when the target speed \(v_t = 2.0 m/s\)}
\begin{tabular}{|l|c|c|}
\hline
 & Proposed method & Comparative method \\
\hline
Completion time (s) & 165.25 & 185.51 \\
\hline
Average linear speed (m/s) & 1.70 & 1.51 \\
\hline
Average angular speed (rad/s) & 0.046 & 0.037 \\
\hline
MAE of lateral (m) & 0.68 & 0.99 \\
\hline
MAE of orientation (rad) & 0.20 & 0.29 \\
\hline
RMSE of linear speed (m/s) & 0.51 & 0.70 \\
\hline
Linear speed deviation (\%) & 15.00 & 23.50 \\
\hline
Accumulated orientation (rad) & 28.64 & 31.61 \\
\hline
\end{tabular}
\end{table}

\chapter{Conclusion}
This study has integrate a vision-based autonomous navigation strategy for AMR operating in race tracks, without relying on HD maps or GNSS. The research was motivated by the limitations of conventional map-based navigation methods in dynamic, unstructured, or GPS-denied scenarios, aiming to develop a lightweight, robust, and real-time navigation framework suitable for practical deployment.

The proposed system is composed of three key modules: perception, path generation, and control. For perception, a deep learning-based model, YOLOP, was employed to extract lane line information from stereo camera inputs. To generate a navigable trajectory, polynomial fitting was applied to the detected lane lines to compute a smooth centerline in the robot’s local coordinate system. A set of three look-ahead points was then selected from the fitted path to serve as references for the control module. The control strategy was based on Lyapunov stability theory, enabling robust and smooth trajectory tracking while accounting for system constraints on linear and angular velocities.

Extensive real-world experiments were conducted using a Honda-provided AI Formula unmanned vehicle on a Formula One race track. The experiments included a variety of scenarios such as straight and curved paths, solid and dotted lane lines, and visually challenging environments like zebra crossings. The results demonstrated that the perception and fitting modules could generally provide accurate and stable inputs to the controller, especially within the critical 2–3 meter region ahead of the vehicle. The Lyapunov-based controller showed superior performance compared to a comparative method in terms of trajectory smoothness, error reduction, and velocity stability, particularly under low speed conditions.

This study evaluates the effectiveness of the proposed control strategy under a preset path tracking framework. In this setting, the robot was required to follow a preset path without relying on real-time perception input. Experimental results demonstrated that the proposed Lyapunov-based controller maintained reliable and smooth trajectory tracking throughout the entire course, even at higher target speeds. In contrast, the comparative method frequently deviated from the preset path, and in some cases, failed to complete the full trajectory due to control instability.

In addition to the preset path tracking experiments, this study also presents the effectiveness of the proposed control strategy under a vision-based tracking framework. Comparisons confirmed that the proposed method achieved lower MAE in lateral and orientation tracking, reduced velocity fluctuations, and more stable angular motion. The proposed method consistently outperformed the baseline in terms of tracking accuracy, velocity consistency, and heading stability, confirming its robustness and adaptability not only in preset path tracking but also in scenarios perception-driven navigation.However, it was also observed that system performance declined slightly at higher target speeds, due to increased sensitivity to visual disturbances and dynamic response limitations. For instance, under the vision-based navigation framework, when the target speed increased from \(1.5\,\text{m/s}\) to \(2.0\,\text{m/s}\), the mean absolute lateral error increased from \(0.20\,\text{m}\) to \(0.26\,\text{m}\), and the mean absolute orientation error rose from \(0.069\,\text{rad}\) to \(0.079\,\text{rad}\). Similarly, in the preset path tracking experiments, the accumulated orientation increased from \(7.43\,\text{rad}\) to \(7.83\,\text{rad}\), and the RMSE of linear speed slightly increased from \(0.014\,\text{m/s}\) to \(0.013\,\text{m/s}\), although still remaining at a low level. Despite these challenges, the overall framework remains effective and adaptable, offering a promising alternative for autonomous navigation in real-world scenarios where reliance on external localization infrastructure is infeasible. This further underscores the controller’s versatility and its suitability for a wide range of autonomous navigation applications.

In future work, we plan to incorporate optimization algorithms to dynamically adjust the controller's preset parameters according to different operating speeds. The goal is to enhance the adaptability of the control system, allowing the robot to maintain a stable and accurate tracking performance across a wider range of speed conditions. Moreover, to address more complex real-world scenarios anticipated in future competitions—such as navigating across zebra crossings and avoiding both static and dynamic obstacles—it will be necessary to introduce additional perception capabilities. In particular, advanced deep learning-based object detection and semantic segmentation methods will be explored to enhance the robot’s understanding of its environment. 
    
\cleardoublepage
\pagenumbering{gobble}
\onecolumn

\bibliographystyle{IEEEtran}
\bibliography{Mybib}


\chapter*{Acknowledgments}
First of all, I would like to express my sincerest gratitude to my supervisor, Professor Cao Wenjing. Professor Cao has always led me forward on the academic path with her keen academic insight. Whenever I was confused or encountered a problem, her timely and profound guidance made me suddenly enlightened and solved the problem. Professor Cao's profound attainments in the field of control theory have benefited me a lot. She made precise and forward-looking suggestions for my research in group meetings and seminars, laying a solid foundation for the successful completion of this thesis.  Moreover, thanks to Professor Takashi Suzuki and Professor Edyta Dzieminska for their diligent participation in my thesis review and defense, offering crucial insights that enhanced my work.

Secondly, I would like to express my sincere gratitude to the students in the robotics group of our laboratory. We inspired each other in discussions, worked side by side in experiments, and reaped results in code modification and debugging. It is this unity and collaboration that has enabled the robot field project to proceed smoothly and made our academic exploration more meaningful.

At the same time, I would like to thank all the members of the Control Lab. At every critical point in the research, everyone generously shared their experiences, made valuable suggestions, and encouraged me when I faced difficulties, which kept me motivated to move forward.

Finally, I would like to thank my parents. Your understanding and support are the most steadfast backing on my road to study, and are also the source of my constant pursuit of excellence and progress.

These experiences will become the most valuable assets in my academic journey. In the future, I will take this as an encouragement to continue exploring in the field I love and constantly surpass myself.

\end{document}